\documentclass[11pt]{article}

\usepackage[preprint]{acl}

\usepackage{times}
\usepackage{latexsym}
\usepackage{rotating} %
\usepackage[T1]{fontenc}
\usepackage[utf8]{inputenc}

\usepackage{microtype}

\usepackage{inconsolata}

\usepackage{graphicx}
\usepackage{amsmath}
\usepackage{amssymb}
\usepackage[ruled,vlined]{algorithm2e}
\usepackage{adjustbox}
\usepackage{booktabs}
\usepackage{caption}
\usepackage{enumitem}
\usepackage{float}
\usepackage[T1]{fontenc}
\usepackage{longtable}
\usepackage{linguex}
\usepackage{listings}
\usepackage{multirow}
\usepackage{natbib}
\usepackage{soul}
\usepackage{subcaption}
\usepackage{relsize}
\usepackage{tabularray}
\usepackage{tikz}
\usepackage{todonotes}
\usepackage{xcolor}
\usepackage{comment}

\lstdefinestyle{promptbox}{
  basicstyle=\ttfamily\small,
  frame=single,
  breaklines=true,
  columns=fullflexible,
  tabsize=2
}

\title{Discovering and Causally Validating Emotion-Sensitive Neurons\\ in Large Audio-Language Models
}

\author{
 \textbf{Xiutian Zhao\textsuperscript{1}},
 \textbf{Bj\"orn Schuller\textsuperscript{2}}, 
 \textbf{Berrak Sisman\textsuperscript{1}}
\\
 \textsuperscript{1} Center for Language and Speech Processing (CLSP), Johns Hopkins University, USA \\
 \textsuperscript{2} Group on Language, Audio \& Music (GLAM), Imperial College London, UK
 }
\begin{document}
\maketitle
\begin{abstract}
Emotion is a central dimension of spoken communication, yet,
we still lack a mechanistic account of how modern large audio-language models (LALMs) encode it internally. We present the first neuron-level interpretability study of emotion-sensitive neurons (ESNs) in LALMs and provide causal evidence supporting the existence of such units in Qwen2.5-Omni, Kimi-Audio, and Audio Flamingo 3. Across these three widely used open-source models, we compare frequency-, entropy-, mean-deviation-, and contrast-based neuron selectors on multiple emotion recognition benchmarks. Using inference-time interventions, we reveal a consistent emotion-specific signature: deactivating neurons selected for a given emotion disproportionately degrades recognition of that emotion while largely preserving other classes, whereas targeted steering amplifies these units to bias predictions toward the target emotion. 
These effects arise with modest amounts of identification data and scale systematically with intervention strength. We further observe that ESNs exhibit non-uniform layer-wise clustering with partial cross-dataset transfer. 
Taken together, our results offer a causal, neuron-level account of emotion decisions in LALMs and highlight targeted neuron interventions as an actionable handle for controllable affective behaviors.
\end{abstract}

\section{Introduction}

The progress of large language models (LLMs) has accelerated the development of multimodal foundation models that jointly process text and other modalities \cite{wang2023large}. Among them, large audio-language models (LALMs) \cite{liu2025voxtral, xu2025qwen25omnitechnicalreport, kimiteam2025kimiaudiotechnicalreport, ghosh2025audio}, which operate on both speech and text, are increasingly prominent in applications such as conversational assistants, where affective competence is crucial for user trust and safety. Despite strong empirical performance, however, we still lack a mechanistic understanding of how LALMs internally represent emotion and which components are responsible for emotion-related decisions  \cite{gandhi2023understanding}.

Speech conveys not only linguistic content but also paralinguistic cues (e.g., intonation, pitch, energy) associated with a speaker's affective state. While decades of affective speech research \cite{wani2021comprehensive} have demonstrated the importance of these cues for tasks such as speech emotion recognition \cite{akccay2020speech, ma-etal-2024-emotion2vec, 10447060, naini25_interspeech} and expressive speech synthesis \cite{9420276, 10003644}, it remains unclear whether LALMs encode emotion through compact, intervention-sensitive neuron sets or through diffuse, non-specific mechanisms.

Neuron-level interpretability provides a natural lens for answering this question. Prior work has demonstrated that individual units can specialize to human-interpretable concepts in vision models \cite{bau2017network, bau2020understanding} and exhibit selectivity for linguistic and other conceptual attributes in LLMs \cite{voita-etal-2024-neurons, yu-ananiadou-2024-neuron, alkhamissi-etal-2025-llm}. In multimodal settings, however, existing neuron-level studies have largely focused on modality- or task-specific patterns rather than affect, and causal validation remains limited \cite{wu2024andaudionetworkdissection, huang2024minerminingunderlyingpattern, fang2024towards, xu2025deciphering}.
Motivated by these gaps, we ask whether LALMs contain compact neuron subsets that function as emotion-sensitive units whose activation can be manipulated to selectively impair or steer emotion-related behavior.

We frame our study around the following questions:

\begin{itemize}[left=0pt]
    \item \textbf{Causality.} Do \emph{emotion-sensitive neurons} (ESNs), i.e., neurons that preferentially activate on inputs tied to particular emotions when processing speech, exist in LALMs? Does deactivating these neurons lead to emotion-specific performance degradation, and can amplifying them systematically steer emotion-related model behavior? 
    \item \textbf{Selectivity.} Which identification methods are most effective at isolating ESNs and are neurons associated with certain emotions intrinsically harder to detect than others?
    \item \textbf{Locality and Transferability.}  How are ESNs distributed across decoder layers, and to what extent do these functional units generalize across datasets and acoustic conditions?
\end{itemize}

Empirically, we study three open-source LALMs that support direct speech input and text generation: Qwen2.5-Omni-7B \cite{xu2025qwen25omnitechnicalreport}, Kimi-Audio \cite{kimiteam2025kimiaudiotechnicalreport}, and Audio Flamingo 3 \cite{ghosh2025audio}. As probe testbeds, we use three established speech emotion recognition (SER) benchmarks: \textsc{IEMOCAP} \cite{busso2008iemocap}, \textsc{MELD} \cite{poria-etal-2019-meld}, and \textsc{MSP-Podcast} \cite{11457331}. To identify ESNs, we compare multiple selectors that operationalize frequency-, entropy-, mean-deviation-, and contrast-based notions of selectivity. Our results provide converging evidence that emotion-sensitive functional units exist in LALMs and can be causally validated.

First, selective deactivation exhibits clear self--cross structure: masking neurons identified for a given emotion disproportionately degrades recognition of that same emotion (\emph{self-deactivation}), while producing substantially smaller average effects on other emotions (\emph{cross-deactivation}).
Critically, we find that not all identification criteria are equally effective: selectors based purely on baseline activation probability or entropy are often insufficient to isolate neurons with more emotion-specific causal effects, whereas mean-deviation- and contrast-based selectors yield neurons with markedly stronger such effects.

Second, we show actionability via activation steering. Amplifying the same neuron sets used for deactivation biases predictions toward the target emotion and can yield positive self--cross gaps, indicating that these units provide a controllable handle rather than being mere correlates.
Beyond label-conditioned (targeted) steering, we also study label-free (agnostic) injection strategies that leverage the discovered ESNs without committing to a chosen source emotion. We evaluate three variants: 2-Pass injection that reinforces the model's initial prediction, Mix injection that softly weight emotion masks using internal evidence, and Union injection that simply boosts the union of all ESNs. Notably, the gap between reliable targeted steering and mixed agnostic outcomes suggests ESNs may interact non-additively under joint amplification.

Third, we analyze where ESNs reside and how well they transfer. We observe non-uniform layer-wise locality patterns, with ESNs clustering in earliest layer (0), early-mid (6--8), and later (19--22) decoder layers rather than being evenly distributed, and we find asymmetric, yet non-trivial cross-dataset transferability across emotions---suggesting partial robustness but also dataset- and category-dependent specificity. 

Overall, this work contributes: (1) to our knowledge, the first neuron-level causal analysis of emotion representations in multiple LALMs via self/cross deactivation and steering across multiple datasets; (2) a systematic comparison of identification methods showing which criteria best isolate causally emotion-sensitive units and the impacts of selecting parameters; and (3) evidence that these units have structured locality, non-trivial cross-dataset transfer, and can be leveraged for both targeted and label-free control of affective behavior in speech-enabled foundation models.

\section{Related Work}
\paragraph{Neuron Specialization and Unit Selectivity.}
Identifying neurons that respond selectively to specific features or concepts is a long-standing theme in interpretability. Prior work has shown that individual units in deep networks can align with human-interpretable concepts.
In vision, Network Dissection quantified such alignments for CNN units \cite{bau2017network,bau2020understanding}, and recent work extends neuron-level interpretability to LLMs, including evidence that some neurons exhibit partial concept selectivity \cite{antverg2022on, huben2024sparse, voita-etal-2024-neurons,yu-ananiadou-2024-neuron,tang-etal-2024-language}, especially through causal tracing style localization \cite{meng2022locating, alkhamissi-etal-2025-llm}.
Most closely related, studies on LLMs investigate affective mechanisms: \citet{lee-etal-2025-large} report clustered emotion neurons with deactivation-based validation, and \citet{wang2025llmsfeelemotioncircuits} identify emotion circuits and demonstrate controllability via steering.

\paragraph{Neuron-Level Interpretability in Multimodal and Audio Models.}
Neuron-level analyses have also been applied to multimodal models, typically to characterize modality- or task-specific attributions \cite{huang2024minerminingunderlyingpattern,fang2024towards,xu2025deciphering}.
In the audio domain, interpretability studies of audio and speech transformers often rely on layer-wise probing or attribution methods to reveal what information (e.g., phonetic, speaker, prosodic cues) is encoded across representations \cite{10031189, Akman2025Improving, 11434723}. Audio Network Dissection \cite{wu2024andaudionetworkdissection} labels acoustic units with natural language descriptors by summarizing responsive audio snippets, but it primarily targets generic acoustic/structural concepts and provides limited evidence for emotion-specific causal roles.

\section{Methods}
\label{sec:method}
We study whether LALMs contain neurons that are selectively active for emotions by coupling activation-based neuron analysis with causal interventions. Concretely, we follow an activation-based log--identify--intervene workflow \cite{huo-etal-2024-mmneuron, huang2024minerminingunderlyingpattern, fang2024towards, tang-etal-2024-language} with SER as a diagnostic task: (1) we instrument decoder MLPs and collect neuron activations while the unintervened model answers correctly, (2) we score and select neurons for emotion sensitivity and construct masks, and (3) we intervene at inference time by manipulating identified neurons and quantifying their causal effects.

\subsection{Activation Logging and Emotion-Sensitive Neuron Identification}
\label{subsec:identification}

We attach forward hooks to the decoder MLP feed-forward blocks and log internal activations on correctly solved SER items. The motivation is pragmatic: restricting to correct items reduces contamination from failure-mode generations and yields cleaner emotion-conditioned activation statistics.

Within each decoder MLP, we record the gating signal from the SwiGLU nonlinearity \cite{shazeer2020glu}. Let $u$ and $v$ denote the two pre-activation streams, and let $g=\mathrm{SiLU}(u)$ be the gated branch that modulates $v$. For layer $l$, neuron index $n$, and token position $t$, we denote the logged scalar gate activation by $a_{l,n,t}$ (the $n$-th coordinate of $g$ at position $t$). These values serve as the basis for all subsequent statistics.

\paragraph{Activation Statistics.} 
Let $\mathcal{E}$ be the emotion set. For each identification example labeled $e\in\mathcal{E}$, we aggregate gate activations across valid token positions, using an indicator $m_{t}\in\{0,1\}$ to exclude padding and other special markers. For every neuron $(l,n)$ we maintain: (1) a positive-activation count $K_{l,n}^{(e)}$, and (2) the total number of valid token positions $T_{e}$ contributed by emotion-$e$ examples:

$$K_{l,n}^{(e)}=\sum_{t}m_{t}\mathbb{I}(a_{l,n,t}^{(e)}>0)$$
$$T_{e} = \sum_{t} m_{t}$$

Intuitively, $K_{l,n}^{(e)}$ captures how often the unit is active under emotion $e$. Normalizing by $T_{e}$ yields the emotion-conditioned activation probability profile, $P_{l,n}^{(e)}$, which serves as the sufficient statistic for all selectors below:

$$P_{l,n}^{(e)} = \frac{K_{l,n}^{(e)}}{T_{e}}$$

Here, $P_{l,n}^{(e)}$ reflects the empirical firing frequency (i.e., probability of positive activation) for the given neuron.

\paragraph{Identification Methods.} Given $P_{l,n}^{(e)}$, we score neurons using the following four established methods, in addition to an emotion-independent random selection baseline.

\begin{itemize}[left=0pt]

    \item \textbf{Random Selection (RND)} is employed as an emotion-agnostic control. We construct a random mask by sampling the same total budget of neurons (i.e., the same $r$ used by targeted selectors) uniformly over all decoder MLP neurons. We do \emph{not} enforce layer-wise matching because it adds bookkeeping and compute overhead; in pilot checks, a layer-matched variant produced similar intervention effects within sampling variance. We therefore report the simpler global RND baseline throughout. Unless stated otherwise, we report RND results averaged over 5 independent random masks of different seeds.

    \item \textbf{Activation Probability (LAP)} (\citealp{huben2024sparse, gurnee2024universal}) prioritizes neurons that are frequently active for a particular emotion, using only the frequency statistic $P^{(e)}_{l,n}$:

    \begin{equation}
      \mathrm{LAP}_{l,n}^{(e)} \;=\; P_{l,n}^{(e)}
      \;=\; \frac{K_{l,n}^{(e)}}{T_{e}}
    \end{equation}

    \item \textbf{Activation Probability Entropy (LAPE)} (\citealp{tang-etal-2024-language, namazifard-poech-2025-isolating}) evaluates selectivity across emotions by forming a normalized distribution over $e\in \mathcal{E}$ for each neuron and computes its Shannon entropy. Lower entropy corresponds to more concentrated firing and thus stronger specialization:

    \begin{align}
    \mathrm{LAPE}_{l,n} &= 
    -\!\sum_{e\in\mathcal{E}} 
    \tilde P_{l,n}^{(e)} \log \tilde P_{l,n}^{(e)},\\
    \tilde P_{l,n}^{(e)} &= 
    \frac{P_{l,n}^{(e)}}{\sum_{e'} P_{l,n}^{(e')}} .
    \end{align}

    Since LAPE assigns each neuron a single selectivity score $\text{LAPE}_{l,n}$ that is not conditioned on any specific emotion, we deterministically map neurons to emotions using the estimated firing probabilities $P^{(e)}_{l,n}$ to enable per-emotion interventions.

     \item \textbf{Mean Activation Difference (MAD)} (\citealp{bau2018identifying, 10.1609/aaai.v33i01.33016309}) evaluates selectivity by measuring how much a neuron's activation probability for a specific emotion deviates from its average activation probability across all emotions. Large absolute values indicate neurons that fire distinctively (either significantly more or significantly less frequently) for emotion $e$ compared to their baseline behavior across all categories:

    \begin{align}
    \mathrm{MAD}_{l,n}^{(e)} &=  \left| P_{l,n}^{(e)} - \overline{P}_{l,n} \right|, \\
    \overline{P}_{l,n} &=  \frac{1}{|\mathcal{E}|} \sum_{e' \in \mathcal{E}} P_{l,n}^{(e')}.
    \end{align}
    
    \item \textbf{Contrastive Activation Margin (ConAct)} (\citealp{zhao-etal-2026-finding}) is a margin-style criterion: for each neuron, it compares the top firing probability across emotions with the runner-up, and assigns the margin to the highest-scoring emotion while assigning zero to all other emotions. 
    Concretely, using the firing probabilities $P_{l,n}^{(e)}$, define:

\begin{align}
\nonumber
P^{(1)}_{l,n} = \max_{e\in\mathcal{E}} P_{l,n}^{(e)}, \quad
e^{(1)}_{l,n} = \arg\max_{e} P_{l,n}^{(e)}\\
P^{(2)}_{l,n} = \max_{e\in\mathcal{E}\setminus\{e^{(1)}_{l,n}\}} P_{l,n}^{(e)}, \\
\mathrm{ConAct}_{l,n}^{(e)} = \begin{cases} P_{l,n}^{(1)} - P_{l,n}^{(2)}, & \text{if } e = e_{l,n}^{(1)} \\ 0, & \text{otherwise.} \end{cases}
\end{align}
\end{itemize}

\paragraph{Emotion-Sensitive Neurons Ranking.}
\label{sec:ESN}

For each selector $m$ and each emotion $e$, we obtain a global ranking of candidate neurons by their emotion-$e$ score. We treat the intervention size as a hyperparameter and, for method comparability, always select a fixed fraction $r$ of the highest-ranked ones as the \emph{emotion-sensitive neurons (ESNs)}.
Formally, let $D_l$ be the width of the monitored MLP gate vector at decoder layer $l$ (i.e., the number of gate units/neuron dimensions we log in that layer). For selector $m$ and emotion $e$, we denote the chosen index set by
$\mathcal{I}^{(m,e)}_l \subseteq \{1,\dots,D_l\}$
and use $\{\mathcal{I}^{(m,e)}_l\}_l$ as the mask support for deactivation and steering in \S 3.2.

\subsection{Intervention: Deactivation, Targeted Steering and Agnostic Injection}
\label{subsec:intervention}

To test whether the identified ESNs are not merely correlational but causally influential in emotion-related decisions, we intervene on their gate activations at inference time. Let $g_{l,t}\in\mathbb{R}^{D_l}$ denote the SwiGLU gate output at decoder layer $l$ and token position $t$, i.e., $g=\mathrm{SiLU}(u)$. Given ESN indices $\mathcal{I}^{(m,e_{\text{src}})}_l$, we build a layer-specific mask that either suppresses or amplifies exactly those indices while leaving all other parameters unchanged.

\paragraph{Deactivation.}
We evaluate necessity by zeroing the selected neurons through an elementwise mask:
\begin{equation}
r^{(m,e_{\mathrm{src}})}_{l,n}=\begin{cases}
0, & n\in \mathcal{I}^{(m,e_{\mathrm{src}})}_l\\[2pt]
1, & \text{otherwise,}
\end{cases}
\label{eq:maskdef}
\end{equation}
and applying it to the gate vector:
\begin{equation}
\tilde g^{\mathrm{abl}}_{l,t}=g_{l,t}\odot r^{(m,e_{\mathrm{src}})}_l.
\label{eq:abl}
\end{equation}

\paragraph{Targeted (emotion-specific) Steering.} 
By scaling the same coordinates with a gain factor $\alpha\ge 0$ using a per-layer scale vector $s_l(\alpha)$ \cite{turner2024steeringlanguagemodelsactivation}, yielding the steered gate $\tilde g^{\text{steer}}_{l,t}$. This intervention increases the contribution of ESN dimensions associated with $e_{\text{src}}$ without modifying weights and is applied to evaluate controllability. 

\begin{equation}
s_{l,n}(\alpha)=\begin{cases}
1+\alpha, & n\in \mathcal{I}^{(m,e_{\mathrm{src}})}_l\\[2pt]
1, & \text{otherwise,}
\end{cases}
\label{eq:scaledef}
\end{equation}
\begin{equation}
\tilde g^{\mathrm{steer}}_{l,t}=g_{l,t}\odot s_l(\alpha).
\label{eq:steer}
\end{equation}

\paragraph{Agnostic Injection.}
Targeted steering requires specifying a source emotion $e_{\text{src}}$ (hence selecting $\mathcal{I}^{(m,e_{\mathrm{src}})}_l$).
In many settings, however, a \emph{label-free} intervention that leverages the discovered ESNs without committing to a chosen emotion is demanded. We therefore implement three agnostic injection strategies, including 2-Pass, Mix, and Union injections.
These strategies are inspired by classic self-training / bootstrapping ideas \cite{yarowsky-1995-unsupervised,mcclosky-etal-2006-effective,NEURIPS2022_639a9a17}, as well as soft routing / mixture weighting mechanisms \cite{shazeer2017}. Implementation details are provided in Appendix~\ref{appendix:agnostic}.

Beyond serving as label-free control baselines, these agnostic strategies also act as a probe of inter-emotion interactions: unlike targeted steering, they may jointly amplify multiple emotion-linked neuron sets (or amplify a mispredicted set), which can induce interference if affective circuits share downstream bottlenecks or exert competing influences on the final decision.

\begin{table*}[ht!]
\centering
\resizebox{\textwidth}{!}{%
\begin{tabular}{@{}llrrrrrrrrrr@{}}
\toprule
                 &                                        & \multicolumn{5}{c}{\textit{Deactivation}}                                                          & \multicolumn{5}{c}{\textit{Targeted Steering}}                                \\ \cmidrule(lr){3-7} \cmidrule(lr){8-12}
LALM             & Acc.$\Delta$                           & RND     & LAP               & LAPE    & MAD               & ConAct                                    & RND     & LAP              & LAPE             & MAD           & ConAct           \\ \midrule
                 & \multicolumn{1}{l|}{Self-Effect}       & 0.32    & $-$7.62           & 1.04    & $-$13.09          & \multicolumn{1}{r|}{\textbf{$-$13.50}} & 0.01    & 0.12             & 0.36             & 2.48          & \textbf{2.73} \\
Qwen2.5-Omni-7B & \multicolumn{1}{l|}{Cross-Effect Avg.} & --      & $-$7.33           & 0.04    & 0.19              & \multicolumn{1}{r|}{\textbf{1.75}}     & --      & \textbf{0.00}    & $-$0.04          & $-$0.25       & $-$0.60 \\
                 & \multicolumn{1}{l|}{Self--Cross Gap}   & --      & $-$0.29           & 1.00    & $-$13.28          & \multicolumn{1}{r|}{\textbf{$-$15.25}} & --      & 0.12             & 0.40             & 2.73          & \textbf{3.33} \\ \midrule
                 & \multicolumn{1}{l|}{Self-Effect}       & $-$0.51 & 0.27              & $-$0.81 & \textbf{$-$13.63} & \multicolumn{1}{r|}{$-$11.65}          & $-$0.19 & $-$0.99          & $-$0.38          & \textbf{2.25} & 1.94          \\
Kimi-Audio    & \multicolumn{1}{l|}{Cross-Effect Avg.} & --      & $-$0.55           & $-$0.92 & $-$1.27           & \multicolumn{1}{r|}{\textbf{0.44}}     & --      & $-$0.78          & \textbf{$-$0.25} & $-$0.66       & $-$0.39       \\
                 & \multicolumn{1}{l|}{Self--Cross Gap}   & --      & 0.83              & 0.10    & \textbf{$-$12.36} & \multicolumn{1}{r|}{$-$12.09}          & --      & $-$0.21          & $-$0.13          & \textbf{2.91} & 2.78          \\ \midrule
                 & \multicolumn{1}{l|}{Self-Effect}       & $-$0.13 & \textbf{$-$34.62} & $-$6.49 & $-$15.17          & \multicolumn{1}{r|}{$-$14.63}          & $-$0.20 & $-$0.18          & 1.06             & 2.97          & \textbf{3.35} \\
Audio Flamingo 3 & \multicolumn{1}{l|}{Cross-Effect Avg.} & --      & $-$35.30          & $-$1.88 & $-$1.96           & \multicolumn{1}{r|}{\textbf{0.70}}     & --      & \textbf{$-$0.04} & $-$0.30          & $-$0.36       & $-$0.72       \\
                 & \multicolumn{1}{l|}{Self--Cross Gap}   & --      & 0.68              & $-$4.61 & $-$13.21          & \multicolumn{1}{r|}{\textbf{$-$15.33}} & --      & $-$0.14          & 1.36             & 3.33 & \textbf{4.07}        \\ \bottomrule
\end{tabular}%
}

\caption{Macro-averaged effects of deactivation (left) and targeted steering (right) across three datasets ($r=0.5\%$), using ESNs produced by five identification methods.
For each method, we report two evaluation settings:
\emph{self-effect} ($e_{\text{src}} = e_{\text{eval}}$) and \emph{cross-effect} (averaged over $e_{\text{src}}\neq e_{\text{eval}}$).
We quantify emotion specificity via the \emph{self--cross gap} (self minus cross).
All entries are \emph{accuracy-changes} relative to the corresponding full model.
Random selection (RND) samples neurons without emotion conditioning and therefore has no self/cross distinction.
Per-dataset breakdowns are provided in Appendix~\ref{appendix:deact_results} and~\ref{appendix:steer_results} (Table~\ref{table:ablate_iemocap}, \ref{table:ablate_meld} and \ref{table:ablate_msp} for deactivation, Table~\ref{table:steer_iemocap}, \ref{table:steer_meld} and \ref{table:steer_msp} for steering).}
\label{table:ablate_results}

\end{table*}

\section{Experiment Setup}
\paragraph{Models and Datasets.}
We study three open-source LALMs that accept speech input and demonstrate strong general audio understanding, including
Qwen2.5-Omni-7B \cite{xu2025qwen25omnitechnicalreport}, Kimi-Audio \cite{kimiteam2025kimiaudiotechnicalreport} and Audio Flamingo 3 \cite{ghosh2025audio}. 
The model versions and dataset splits are listed in Appendix~\ref{appendix:models} (Table~\ref{table:models}, \ref{table:datasets_split}).

We evaluate the models on three widely used SER benchmarks: \textsc{IEMOCAP} \cite{busso2008iemocap}, \textsc{MELD} \cite{poria-etal-2019-meld}, and \textsc{MSP-Podcast} \cite{11457331}.
We focus on overlapping subsets of discrete emotion categories that are consistently annotated across these datasets. For evaluation, we construct \emph{class-balanced} test subsets with a fixed number of utterances per emotion; for identification, we additionally \emph{cap} the pool of correctly answered items per emotion to make selectors comparable across categories.

\paragraph{Evaluation Protocol (Self vs Cross Effects).}
After selecting ESNs for each source emotion $e_{\text{src}}\in\mathcal{E}$ (for every selector in \S3.1), we evaluate their speech emotion-sensitivity by running the model on held-out utterances with and without intervention.
We report two complementary settings. In the \emph{self-effect} condition ($e_{\text{src}}=e_{\text{eval}}$), the intervention targets neurons identified from the same emotion as the evaluated subset. In the \emph{cross-effect} condition ($e_{\text{src}}\neq e_{\text{eval}}$), we reuse an emotion-$e_{\text{src}}$ mask while evaluating on a different emotion subset $e_{\text{eval}}$. For deactivation, we expect performance to decrease; for targeted steering, we expect increases toward $e_{\text{src}}$. Comparing self vs.\ cross effects isolates whether a mask primarily modulates a specific emotion pathway rather than causing broad, non-specific degradation or global changes in affective processing.

\paragraph{Prompting and Decoding.}
All models are evaluated in a controlled multiple-choice SER format using a single instruction template (Appendix~\ref{appendix:prompt}). To reduce known multiple-choice artifacts such as label/position preferences \citep{zheng2024large,zhao-etal-2024-measuring}, we randomize the option-number$\leftrightarrow$emotion mapping for every evaluation item, so a preference for a particular number cannot systematically inflate any single emotion.
We decode deterministically (greedy; temperature $0$ with sampling disabled) and cap generation at 20 tokens, which is sufficient for the required short-form response. Since some instruction-tuned models may still emit extra text, we post-process generations with a lightweight parsing routine described in Appendix~\ref{appendix:normalize}.

\begin{figure*}[ht!]
    \centering
    \begin{subfigure}[b]{0.24\textwidth}
        \centering
        \includegraphics[width=\linewidth]{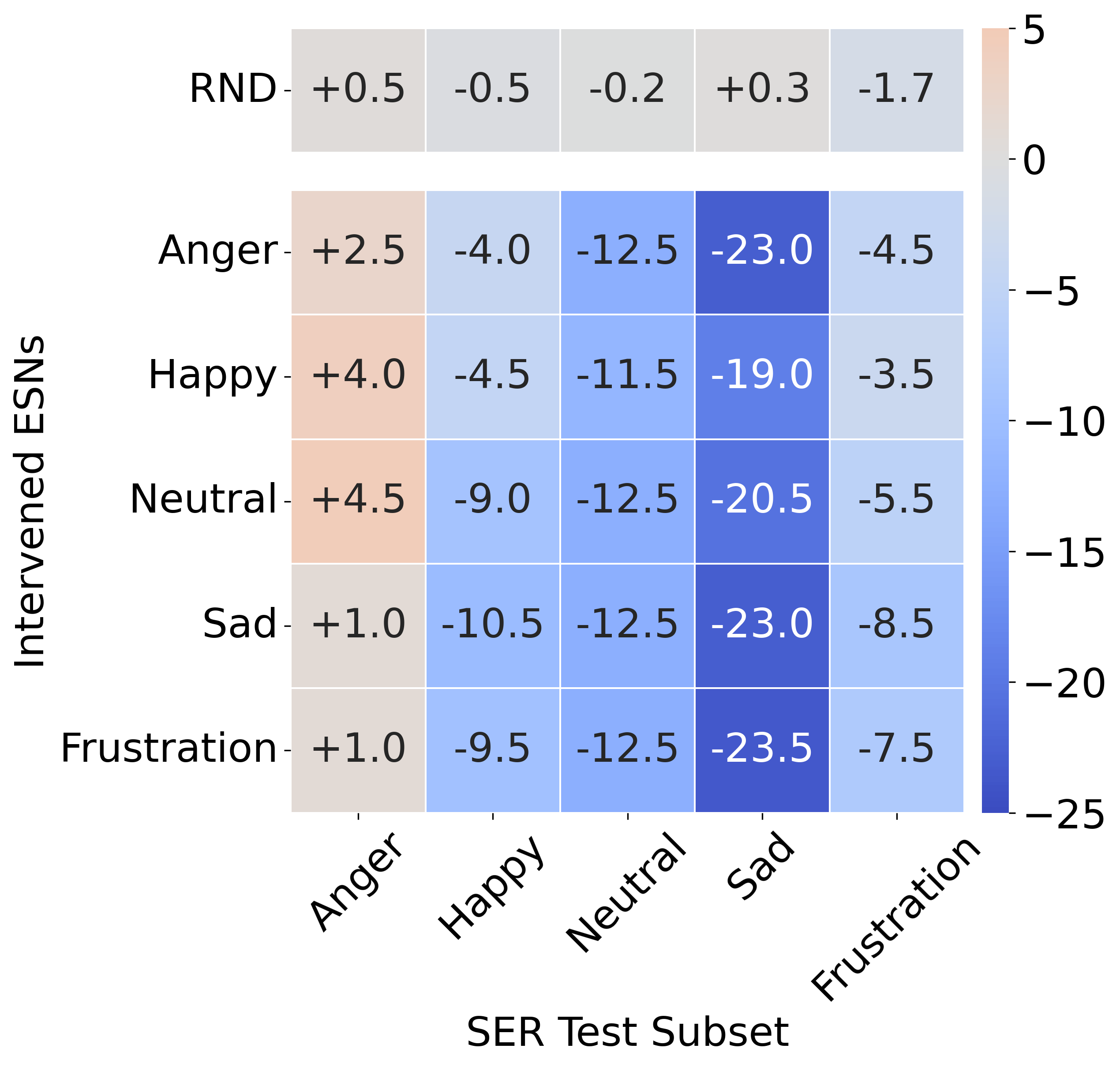}
    \end{subfigure}\hfill
    \begin{subfigure}[b]{0.24\textwidth}
        \centering
        \includegraphics[width=\linewidth]{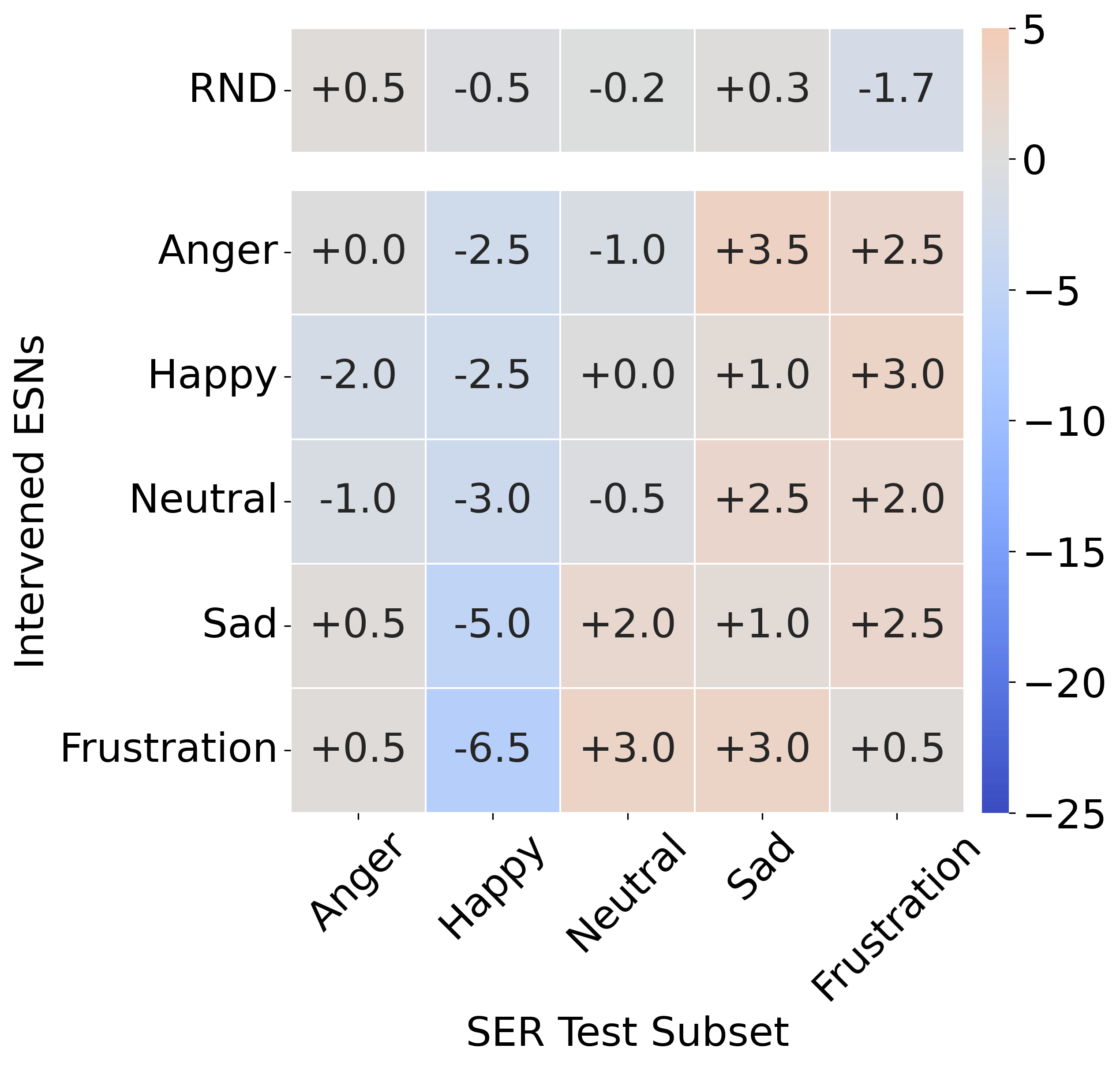}
    \end{subfigure}\hfill
    \begin{subfigure}[b]{0.24\textwidth}
        \centering
        \includegraphics[width=\linewidth]{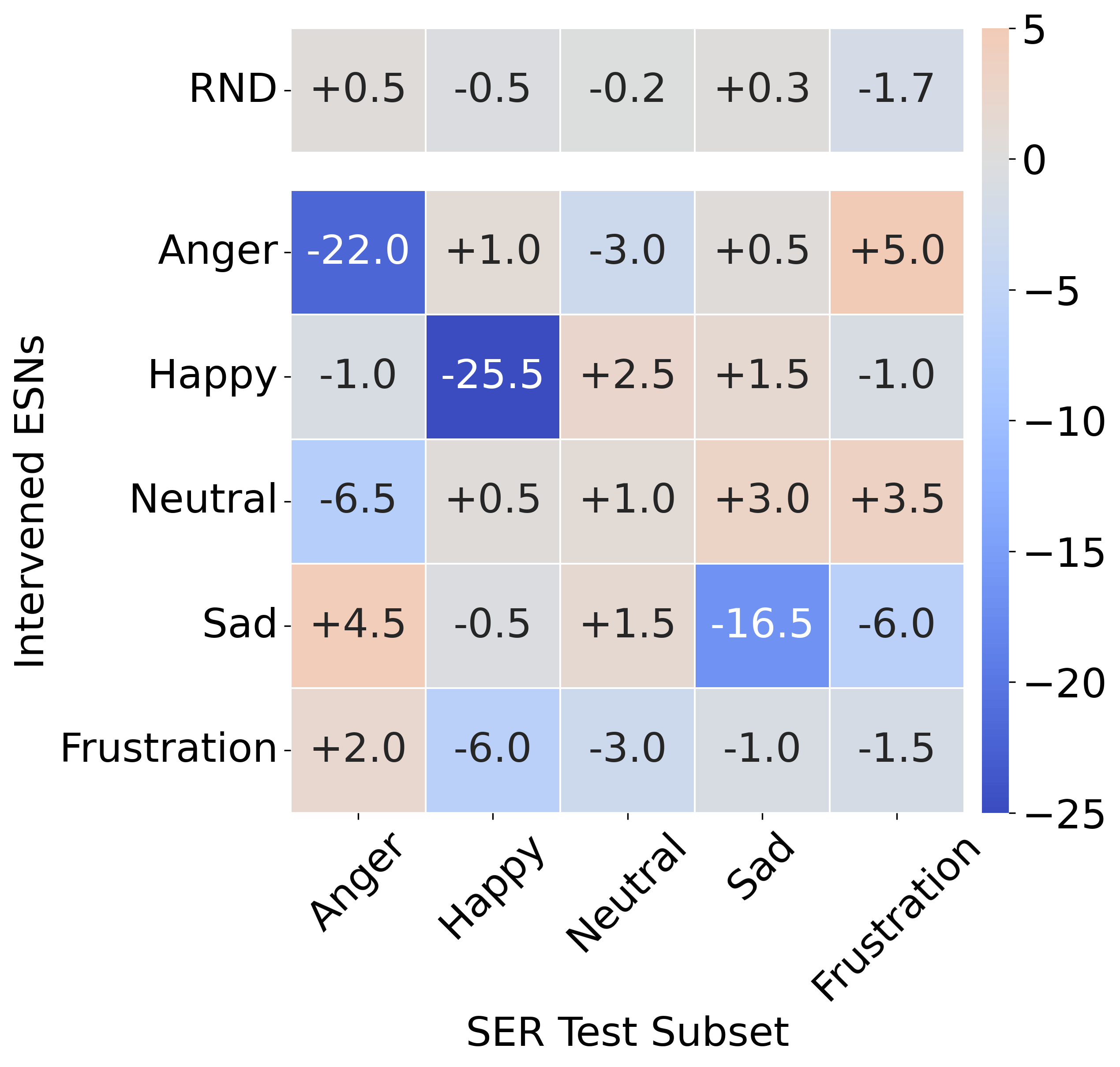}
    \end{subfigure}\hfill
    \begin{subfigure}[b]{0.24\textwidth}
        \centering
        \includegraphics[width=\linewidth]{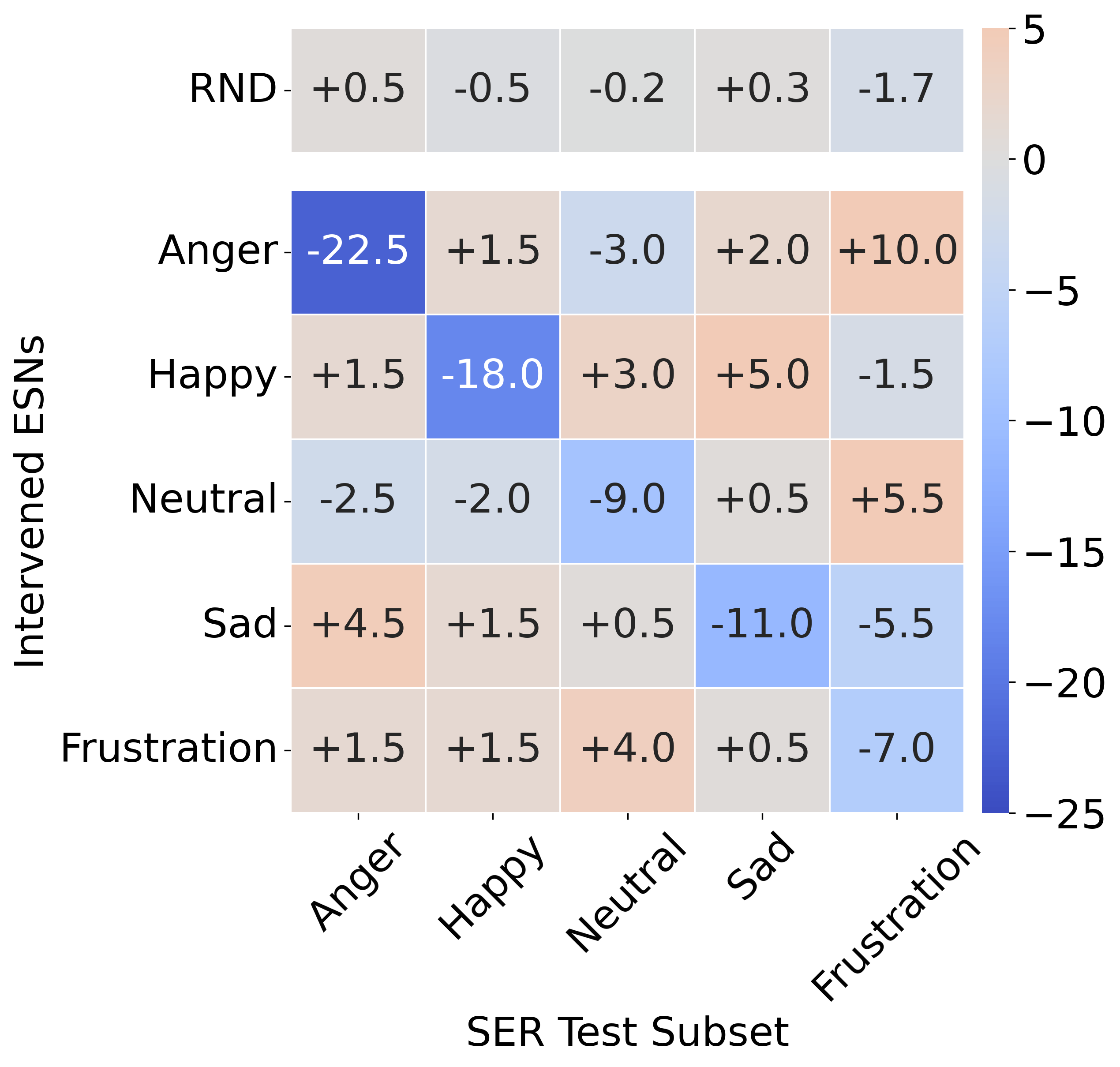}
    \end{subfigure}

    \vspace{0.5em} %

    \begin{subfigure}[b]{0.24\textwidth}
        \centering
        \includegraphics[width=\textwidth]{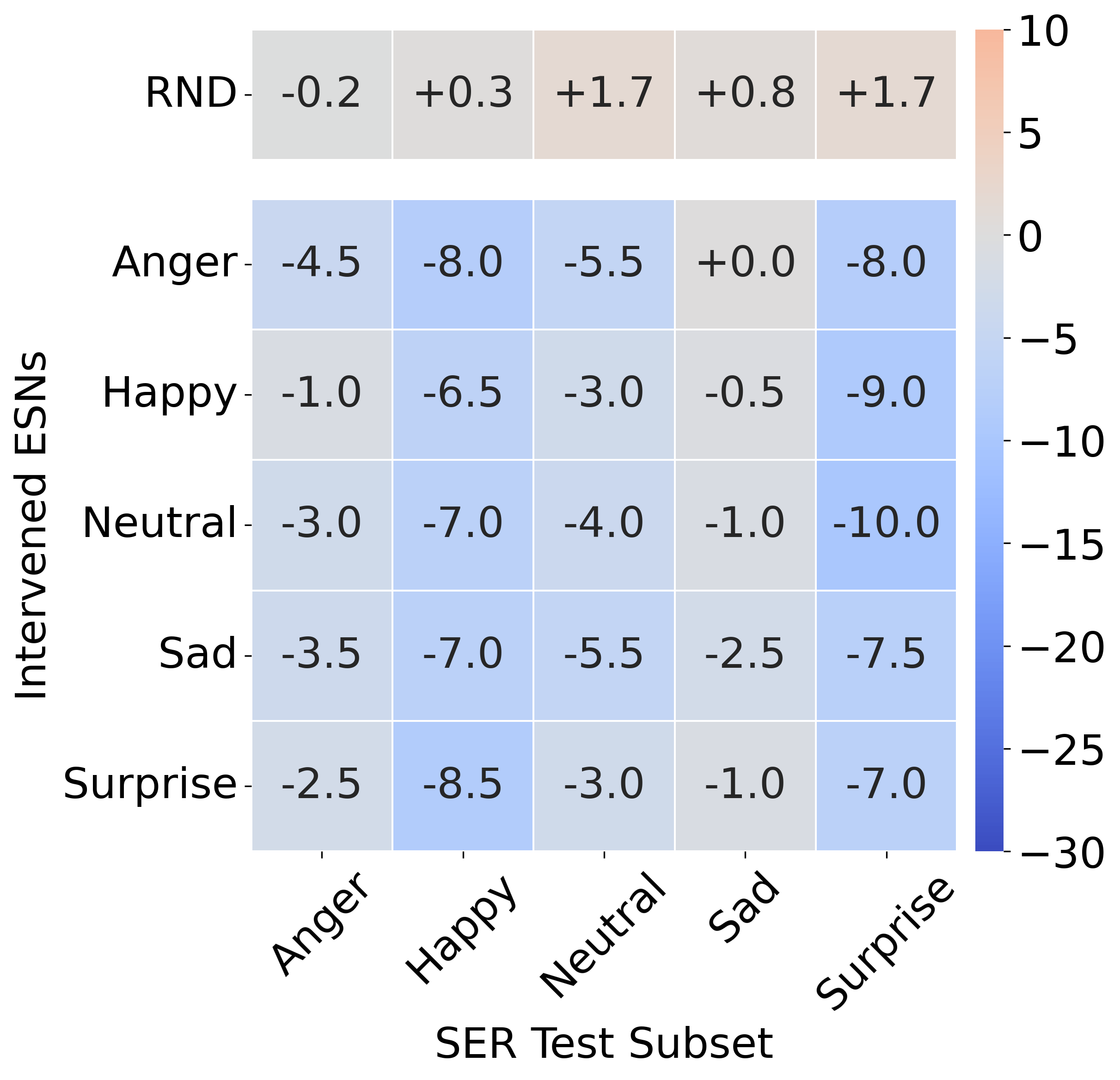}
    \end{subfigure}
    \hfill  
    \begin{subfigure}[b]{0.24\textwidth}
        \centering
        \includegraphics[width=\textwidth]{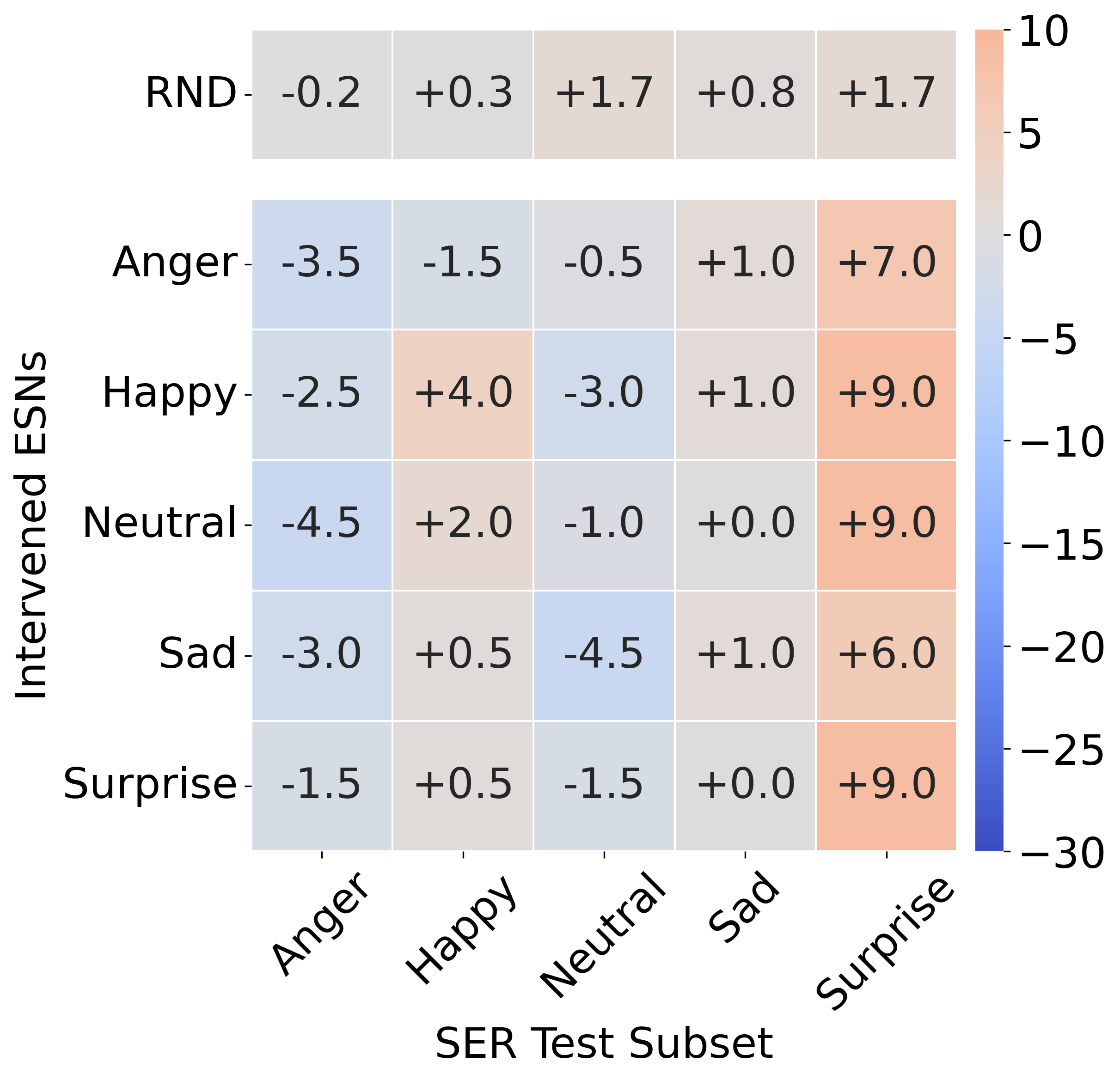}
    \end{subfigure}
    \hfill
    \begin{subfigure}[b]{0.24\textwidth}
        \centering
        \includegraphics[width=\textwidth]{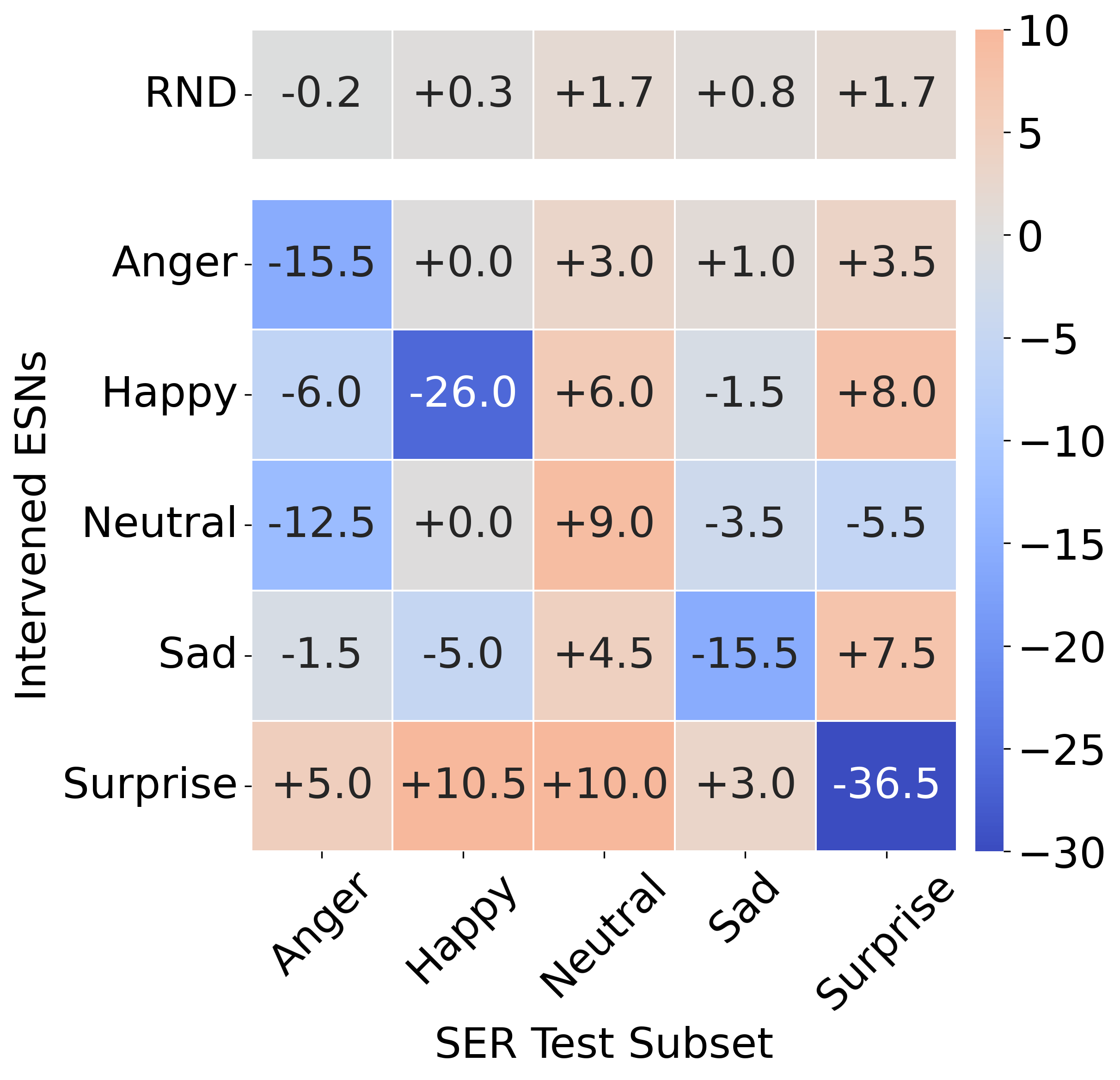}
    \end{subfigure}
    \hfill
    \begin{subfigure}[b]{0.24\textwidth}
        \centering
        \includegraphics[width=\textwidth]{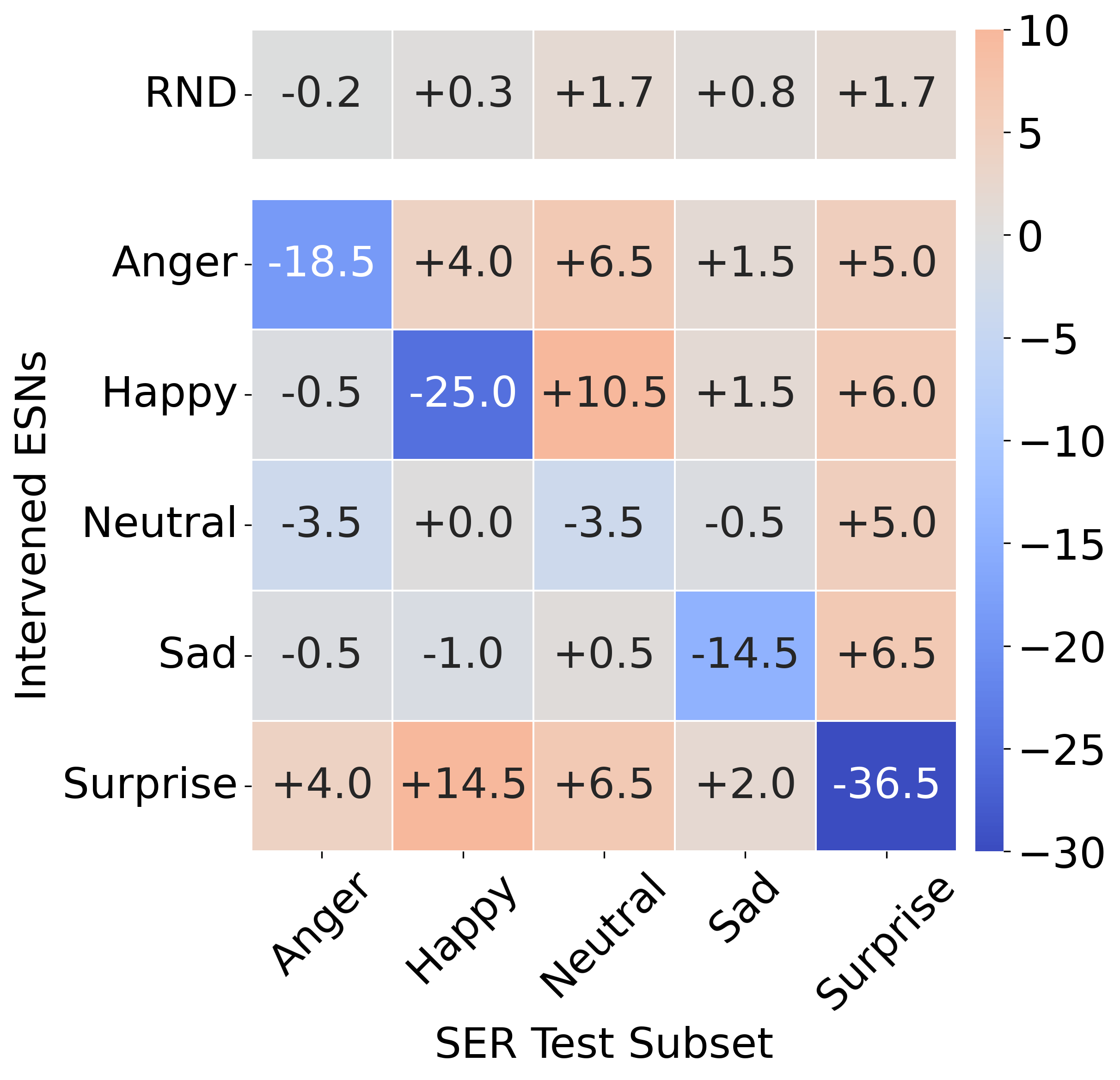}
    \end{subfigure}

        \vspace{0.5em} %

    \begin{subfigure}[b]{0.24\textwidth}
        \centering
        \includegraphics[width=\textwidth]{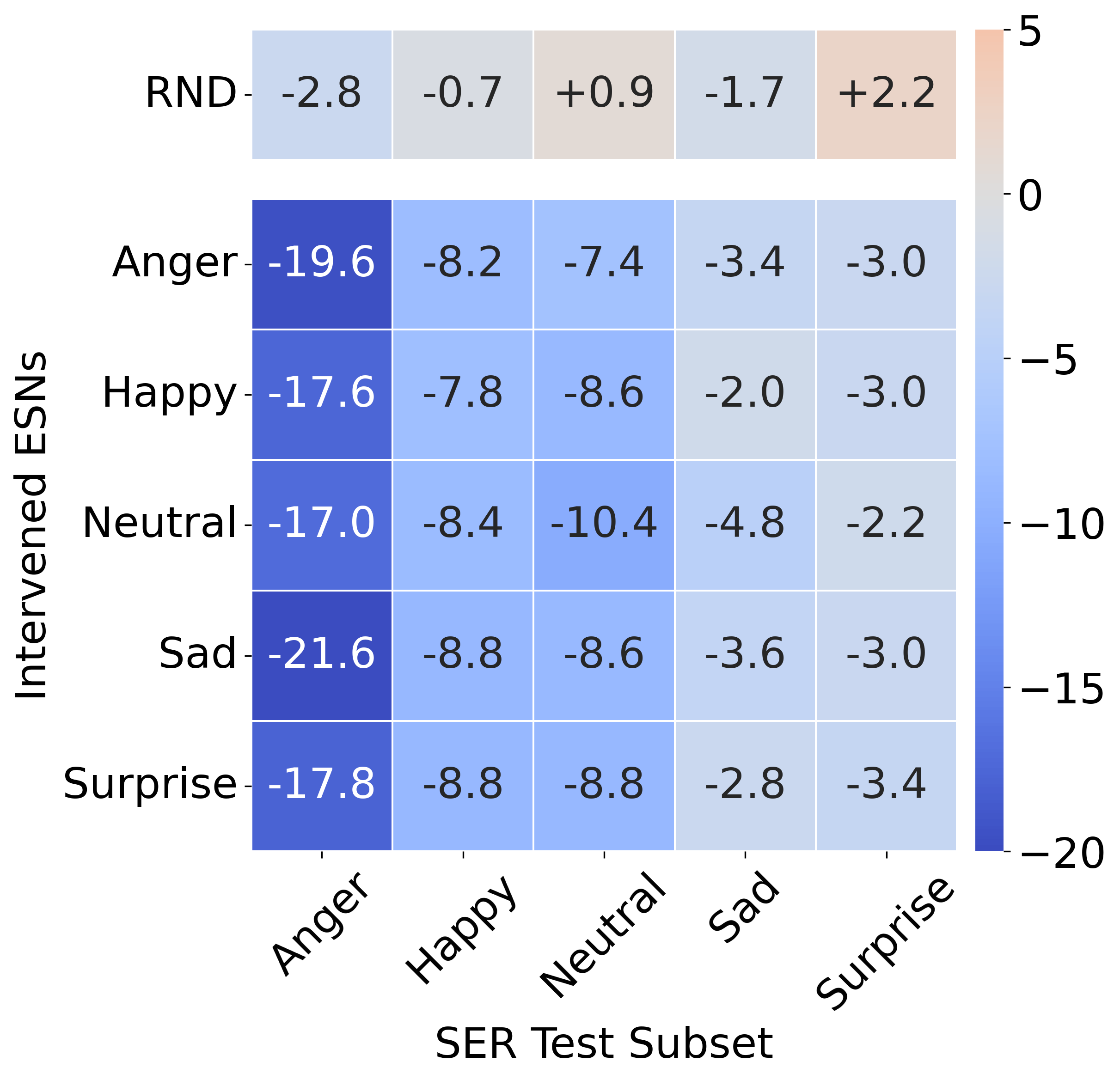}
        \caption{LAP}
    \end{subfigure}
    \hfill  
    \begin{subfigure}[b]{0.24\textwidth}
        \centering
        \includegraphics[width=\textwidth]{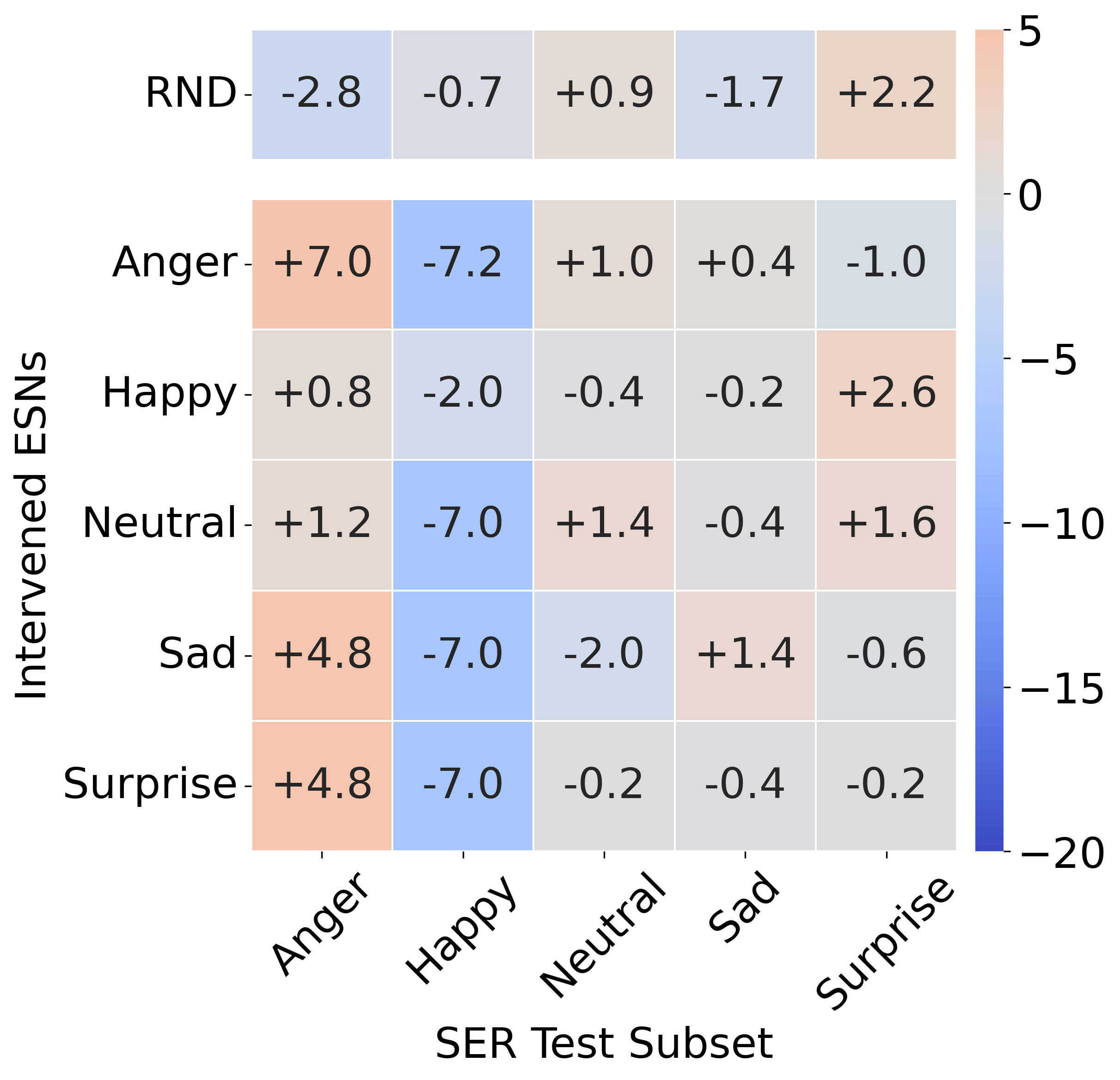}
        \caption{LAPE}
    \end{subfigure}
    \hfill
    \begin{subfigure}[b]{0.24\textwidth}
        \centering
        \includegraphics[width=\textwidth]{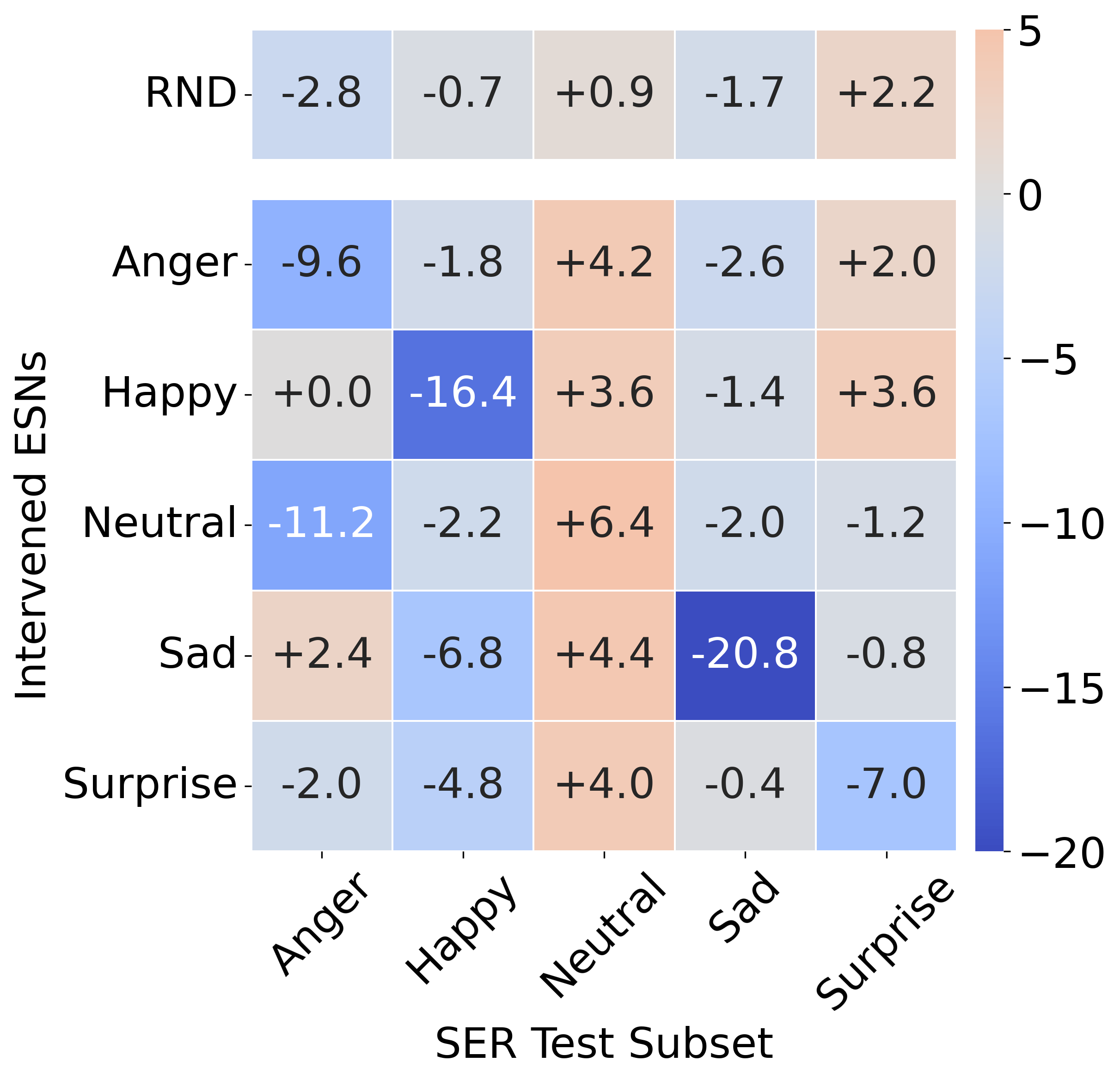}
        \caption{MAD}
    \end{subfigure}
    \hfill
    \begin{subfigure}[b]{0.24\textwidth}
        \centering
        \includegraphics[width=\textwidth]{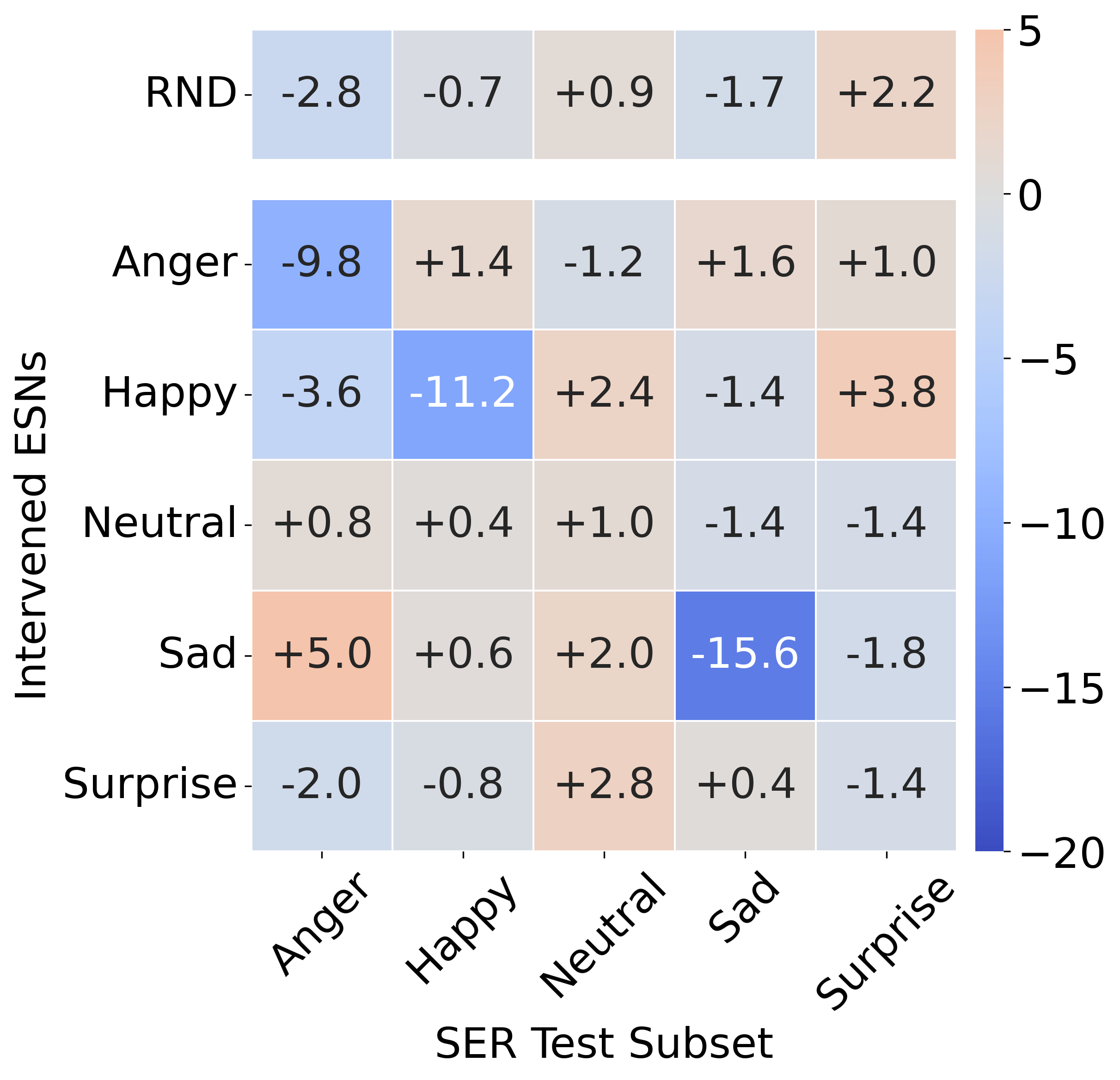}
        \caption{ConAct}
    \end{subfigure}

    \caption{Per-emotion accuracy-change heatmaps for Qwen2.5-Omni-7B under neuron \emph{deactivation}, reported on \textsc{IEMOCAP} (top), \textsc{MELD} (middle), and \textsc{MSP-Podcast} (bottom). Rows index the \emph{source emotion} used to identify the ESN mask; columns index the \emph{evaluation emotion subset}. All values are absolute accuracy differences with respect to the unintervened model. Diagonals correspond to self-effects, while off-diagonals reflect cross-effects.} 
    \label{fig:deact_qwen}
\end{figure*}

\section{Results}

\subsection{Deactivation}
The deactivation section (left half) in Table~\ref{table:ablate_results} shows that masks produced by MAD and ConAct consistently yield a strong separation between self- and cross-effects: performance drops sharply when deactivating ESNs tied to the evaluated emotion, while the average cross-emotion changes are smaller in magnitude, yielding substantial self-–cross gaps.
Across the three LALMs, this manifests as large negative self-effects (11--15 accuracy points) paired with near-zero cross-effects, producing substantial self--cross gaps.
Deactivating the same-sized random masks (RND) produces smaller and less structured changes, supporting that we are not merely removing generic capacity. LAP/LAPE do not reliably produce a clean diagonal signature (often yielding weak, noisy shifts or broader degradation).
Figure~\ref{fig:deact_qwen} visualizes this difference: MAD/ConAct show pronounced diagonal dominance, with limited off-diagonal spillover, consistent with emotion-specific units rather than purely correlated cues.

\begin{figure*}[ht!]
    \centering
    \begin{subfigure}[b]{0.24\textwidth}
        \centering
        \includegraphics[width=\linewidth]{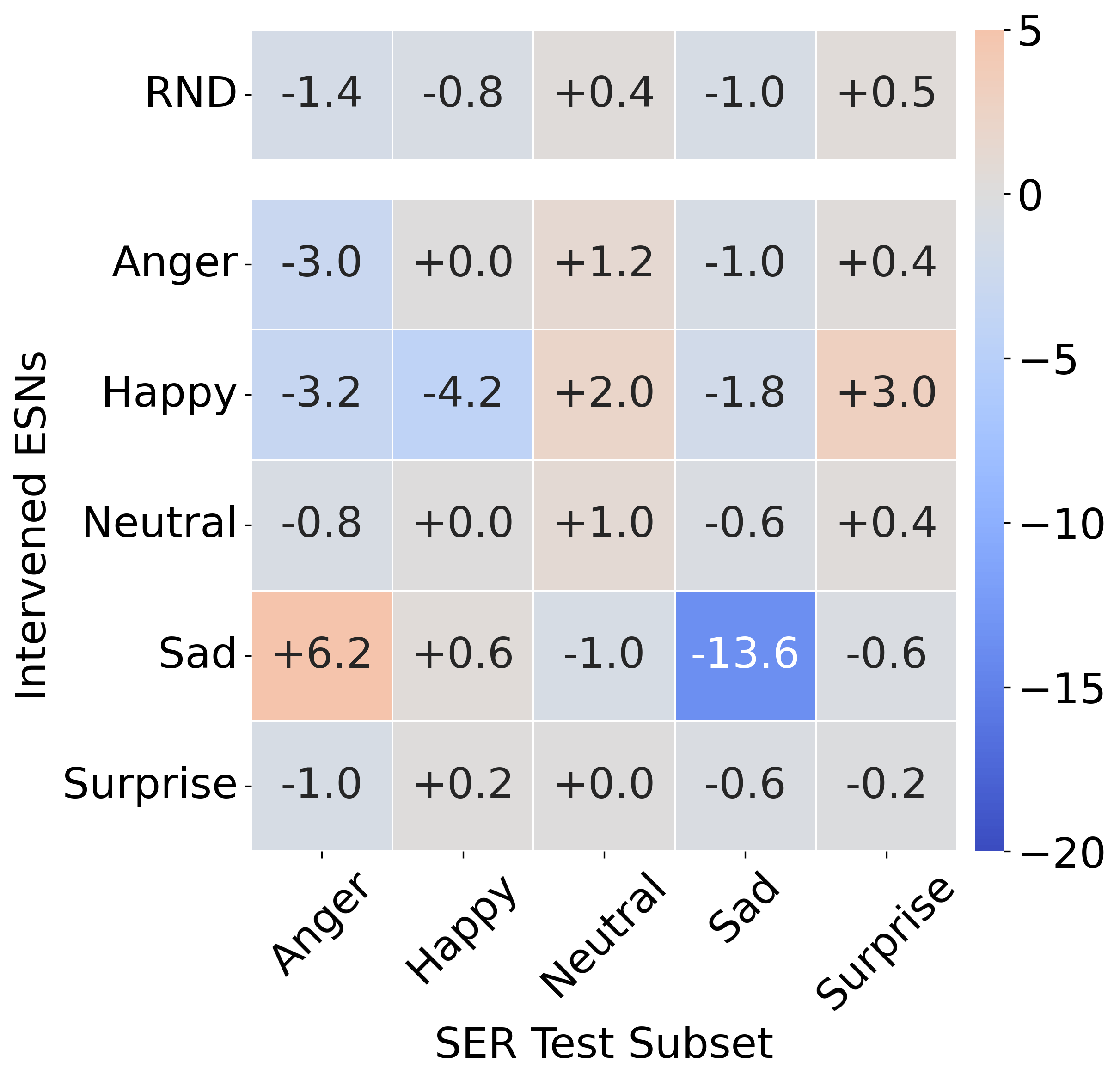}
        \caption{$r=0.1\%$}
    \end{subfigure}\hfill
    \begin{subfigure}[b]{0.24\textwidth}
        \centering
        \includegraphics[width=\linewidth]{figures/acc_delta_qwen25_omni_7b_MSP-PODCAST-Publish-1.12_MSP-PODCAST-Publish-1.12_CAS_ablate_0.005_0.95_1000.png}
        \caption{$r=0.5\%$}
    \end{subfigure}\hfill
    \begin{subfigure}[b]{0.24\textwidth}
        \centering
        \includegraphics[width=\linewidth]{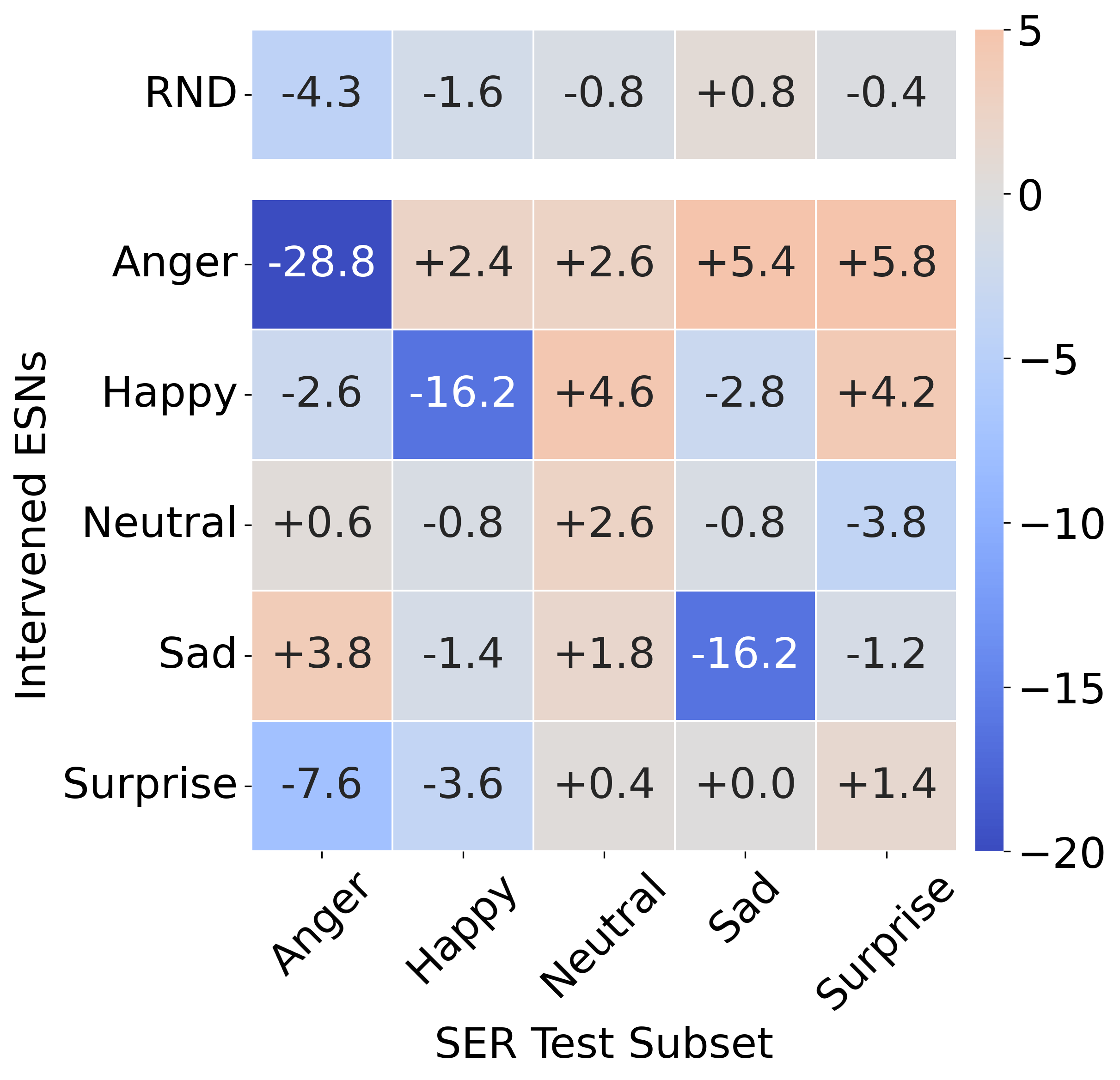}
        \caption{$r=1.0\%$}
    \end{subfigure}\hfill
    \begin{subfigure}[b]{0.27\textwidth}
        \centering
        \includegraphics[width=\linewidth]{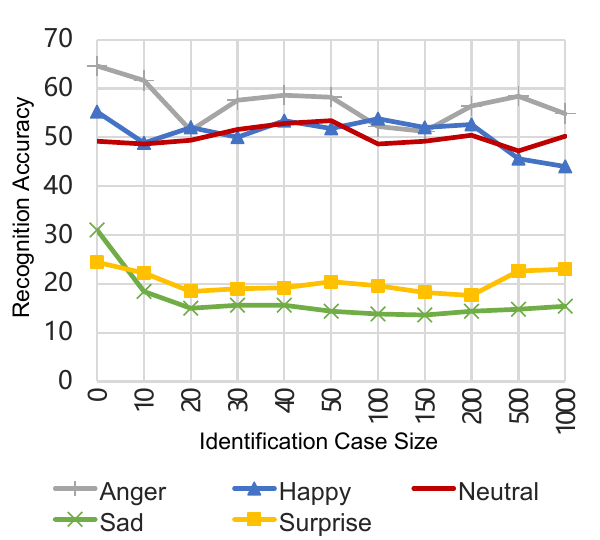}
        \caption{}
    \end{subfigure}

    \caption{Sensitivity to intervention budget and identification-pool size. (a--c) Accuracy-change heatmaps as we vary the deactivated fraction $r$ of ESNs. (d) Accuracies as we vary the number of correctly answered identification examples per emotion used to construct the ESN masks (Qwen2.5-Omni-7B, ConAct-selected ESNs, MSP-Podcast).
    }
    \label{fig:rvalue_example}

\end{figure*}

\paragraph{Effect of ESN Set Size.}
Figure~\ref{fig:rvalue_example}(a--c) varies the fraction of deactivated neurons ($r$) while fixing the model and selector, revealing a trade-off between selectivity and intervention strength. For small $r$, deactivation already yields clear diagonal patterns, indicating that a small subset of neurons can suffice to induce emotion-specific degradation. As $r$ increases, self-deactivation effects strengthen, but off-diagonal changes also grow, reflecting increased collateral disruption of shared circuitry and a shift toward broader capacity loss. Thus, larger masks amplify intervention strength but reduce interpretability by mixing emotion-specific and emotion-general effects. This motivates using a moderate $r$ (i.e., 0.5\%) in subsequent experiments to balance causal potency with clean self--cross dissociation.

\begin{figure*}[ht!]
    \centering
    \begin{subfigure}[b]{0.24\textwidth}
        \centering
        \includegraphics[width=\linewidth]{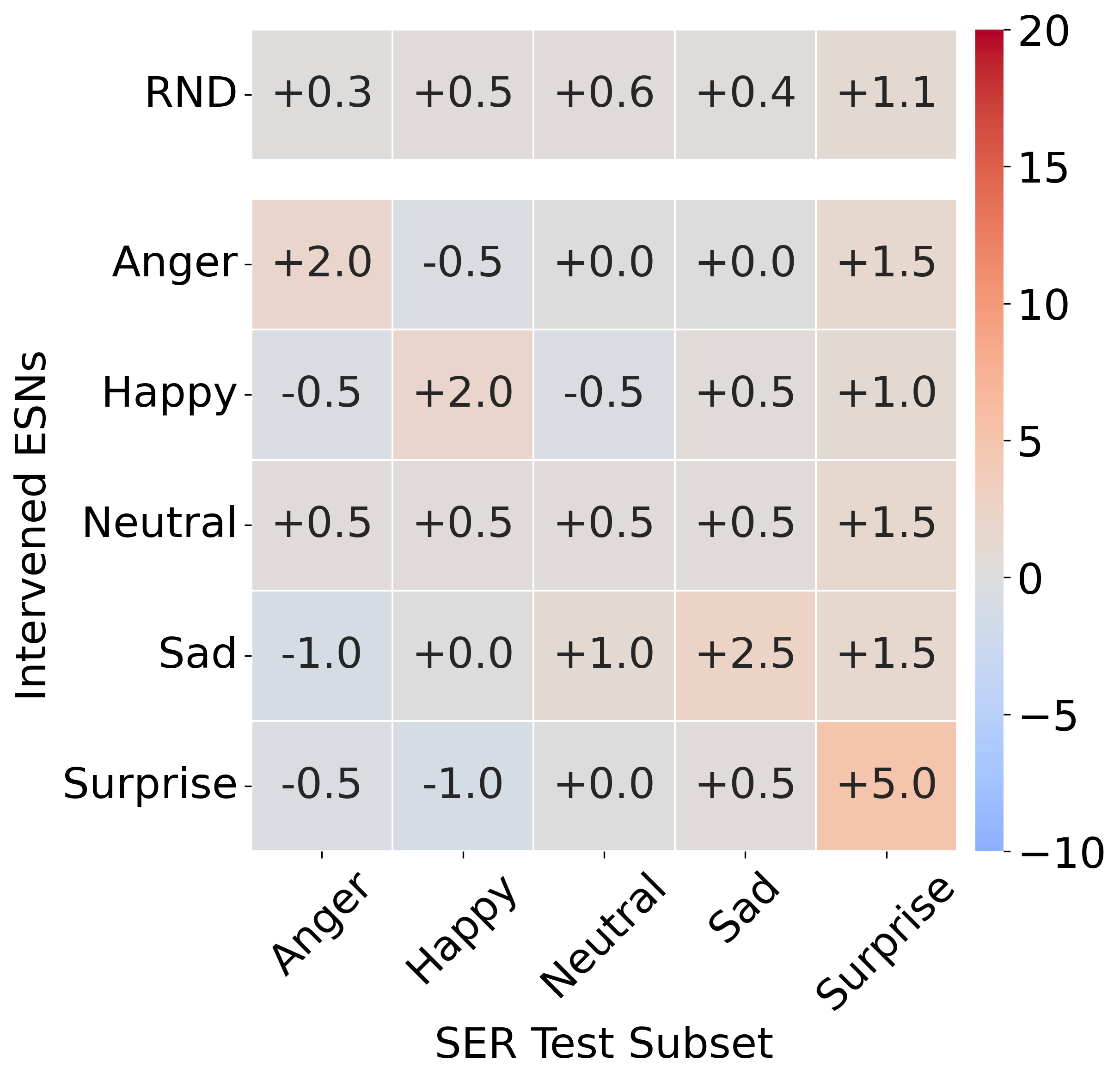}
        \caption{$\alpha=0.10$}
    \end{subfigure}\hfill
    \begin{subfigure}[b]{0.24\textwidth}
        \centering
        \includegraphics[width=\linewidth]{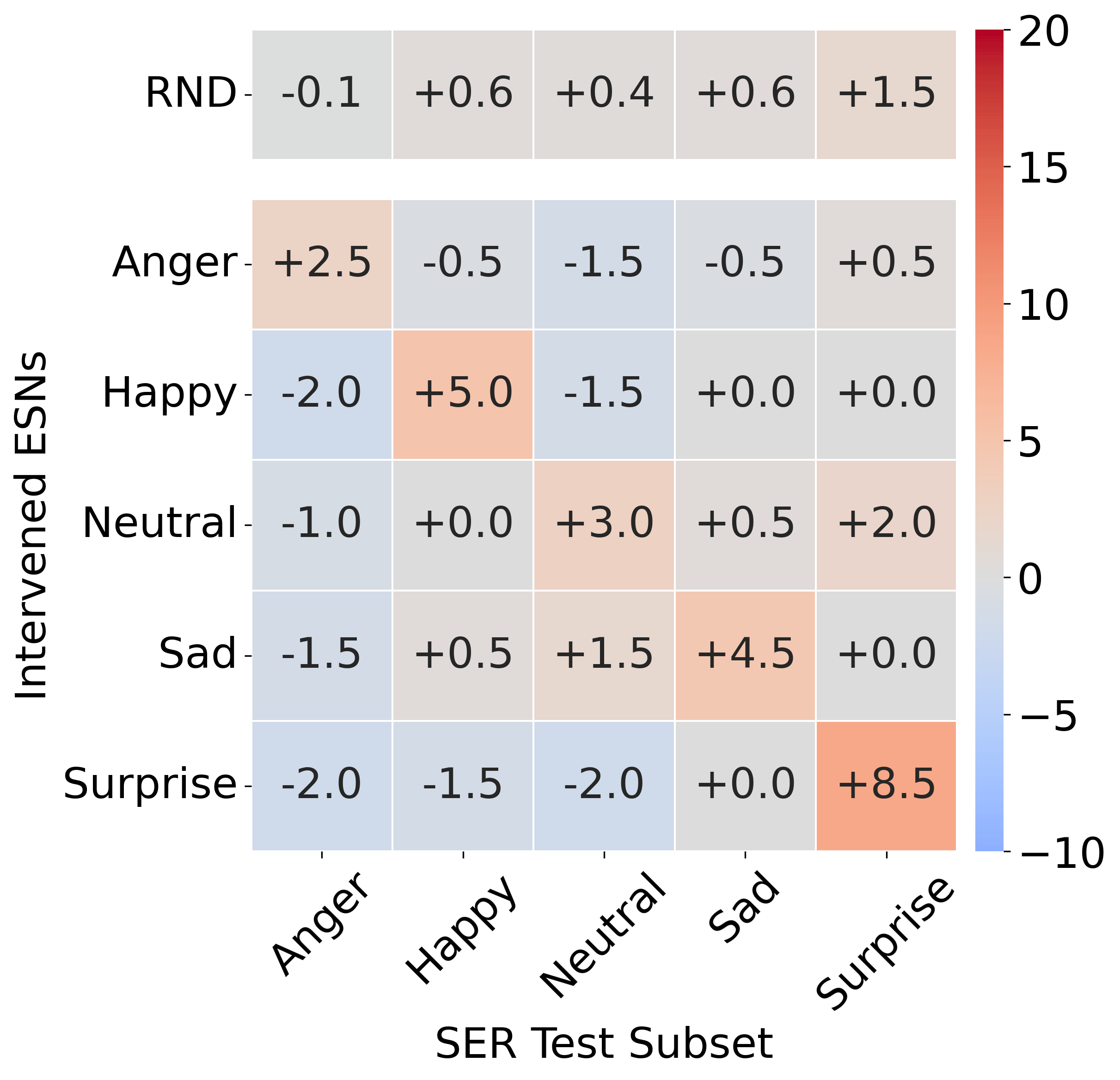}
        \caption{$\alpha=0.30$}
    \end{subfigure}\hfill
    \begin{subfigure}[b]{0.24\textwidth}
        \centering
        \includegraphics[width=\linewidth]{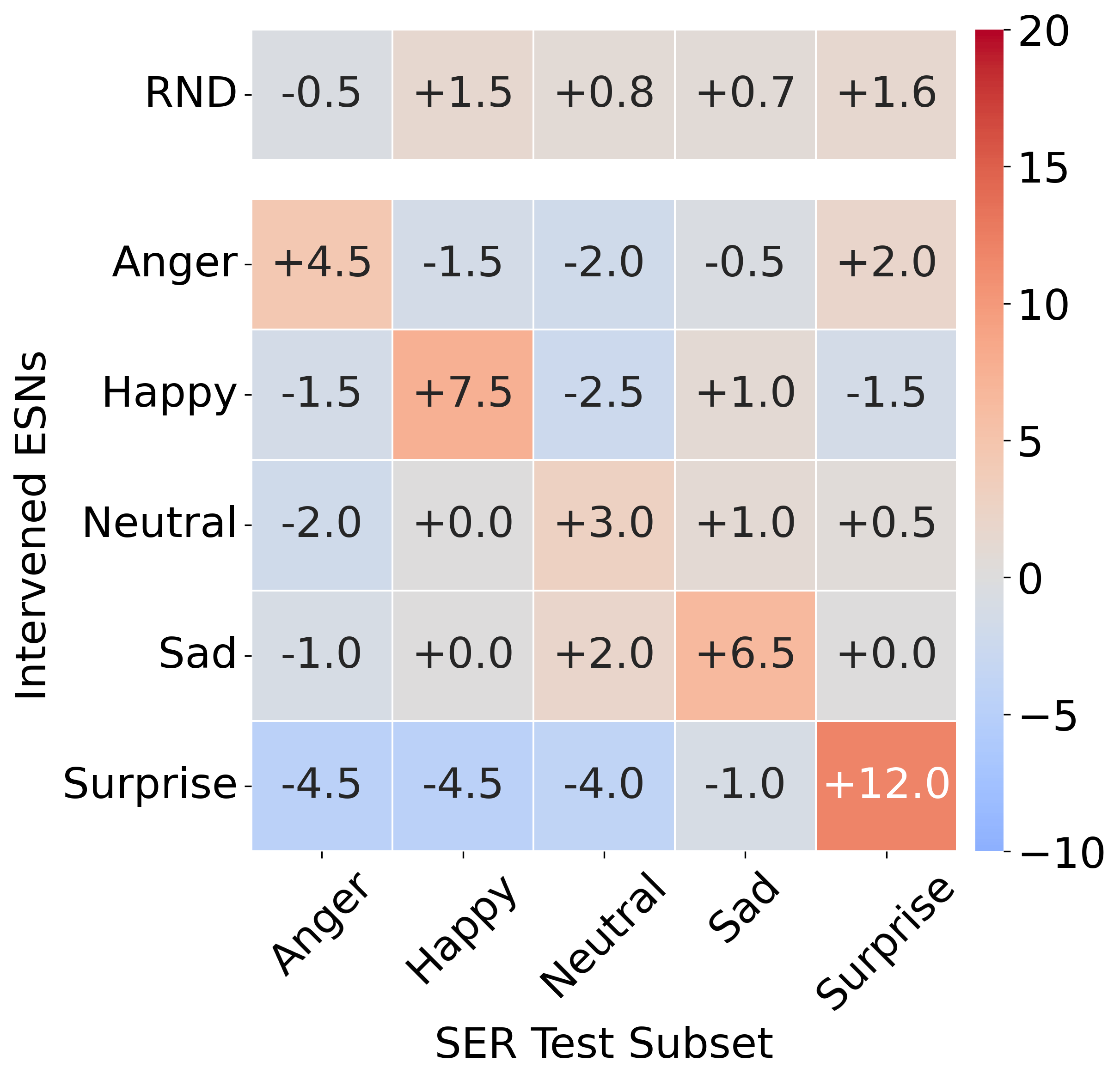}
        \caption{$\alpha=0.50$}
    \end{subfigure}\hfill
    \begin{subfigure}[b]{0.24\textwidth}
        \centering
        \includegraphics[width=\linewidth]{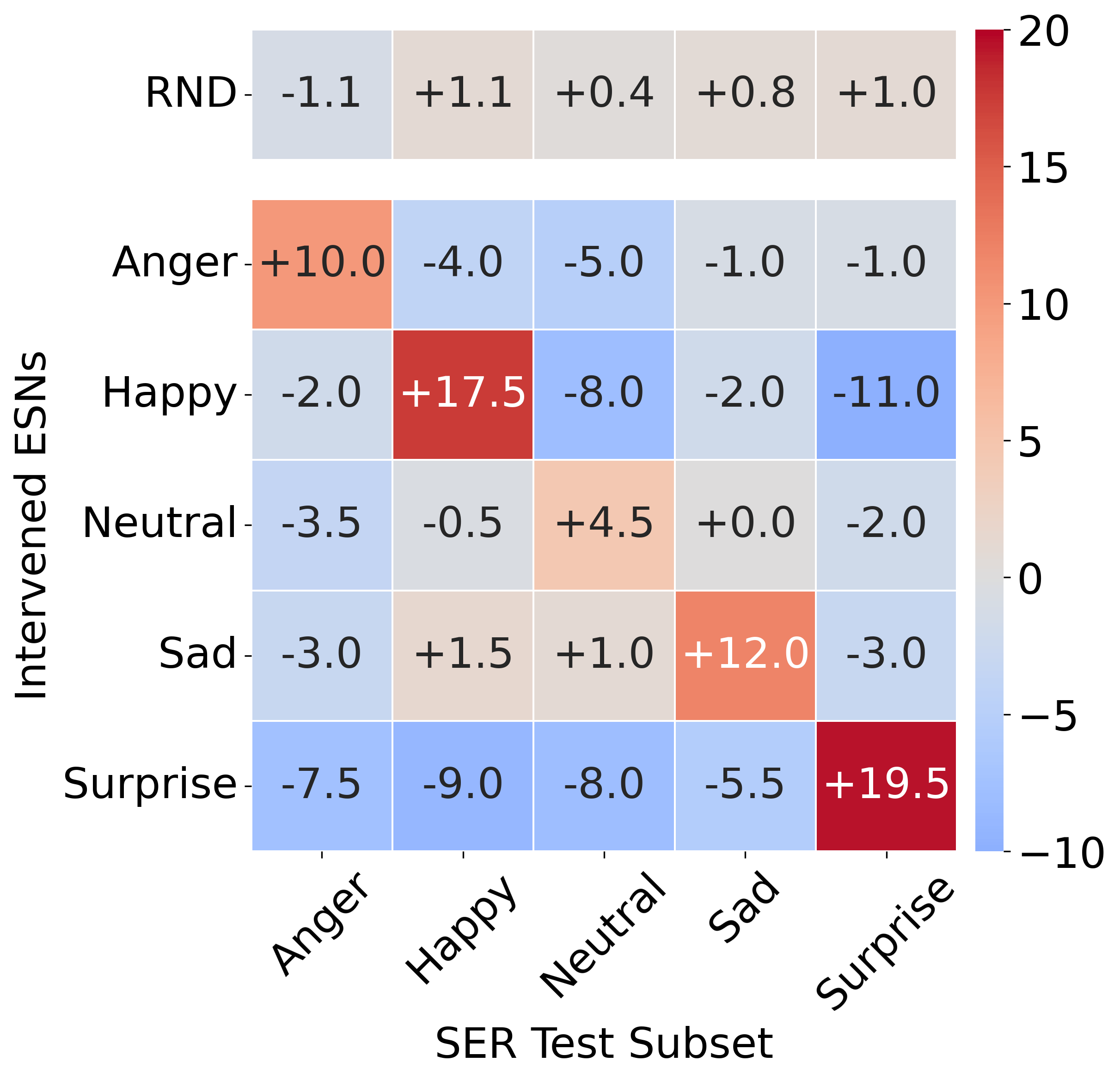}
        \caption{$\alpha=1.00$}
    \end{subfigure}

    \caption{Accuracy-change $\Delta$ heatmaps on \textsc{MELD} for different \textbf{steering strengths} $\alpha$ (ConAct, Qwen2.5-Omni-7B).}
    \label{fig:steer_qwen}

\end{figure*}

\begin{table*}[ht!]
\centering
\resizebox{\textwidth}{!}{%
\begin{tabular}{@{}ll|c|cccc|cccc|cccc@{}}
\toprule
\multirow{2}{*}{Dataset} & \multirow{2}{*}{Model} & \multicolumn{1}{c|}{}         & \multicolumn{4}{c|}{$\alpha=0.1$}                                                                           & \multicolumn{4}{c|}{$\alpha=0.3$}                                                                           & \multicolumn{4}{c}{$\alpha=1.0$}                                                                           \\ \cmidrule(l){4-15} 
                         &                        & \multicolumn{1}{r|}{Unmasked} & \multicolumn{1}{r}{RND} & \multicolumn{1}{r}{2-Pass} & \multicolumn{1}{r}{Mix} & \multicolumn{1}{r|}{Union} & \multicolumn{1}{r}{RND} & \multicolumn{1}{r}{2-Pass} & \multicolumn{1}{r}{Mix} & \multicolumn{1}{r|}{Union} & \multicolumn{1}{r}{RND} & \multicolumn{1}{r}{2-Pass} & \multicolumn{1}{r}{Mix} & \multicolumn{1}{r}{Union} \\ \midrule
                         & Qwen2.5-Omni-7B       & 48.4                          & 48.2                    & 48.1                       & 48.2                    & 48.3                       & 47.9                    & 48.7                       & 48.4                    & 48.9                       & 47.6                    & 49.1                       & \textbf{49.2}           & 49.0                      \\
IEMOCAP                  & Kimi-Audio          & 64.6                          & 64.3                    & 62.0                       & 64.2                    & 64.1                       & \textbf{66.2}           & 61.7                       & 64.2                    & 64.5                       & 63.4                    & 60.6                       & 65.1                    & 60.7                      \\
                         & Audio Flamingo 3       & 59.5                          & 59.7                    & 59.8                       & 59.9                    & 60.0                       & 59.5                    & 59.6                       & 60.2                    & 60.2                       & 59.4                    & 59.3                       & 60.4                    & \textbf{61.6}             \\ \midrule
                         & Qwen2.5-Omni-7B       & 45.3                          & 45.9                    & 45.9                       & 46.1                    & 46.0                       & 45.9                    & 45.7                       & 45.6                    & 46.4                       & 45.7                    & 45.5                       & \textbf{46.7}           & 44.5                      \\
MELD                     & Kimi-Audio          & 57.4                          & 57.3                    & \textbf{58.2}              & 57.5                    & 57.1                       & 57.1                    & 58.0                       & 57.5                    & 56.8                       & 57.1                    & 56.3                       & 56.8                    & 53.2                      \\
                         & Audio Flamingo 3       & 49.4                          & 49.1                    & 49.3                       & 49.6                    & 49.2                       & 49.0                    & 49.3                       & 49.1                    & 49.6                       & 48.7                    & 49.1                       & \textbf{49.7}           & 48.3                      \\ \midrule
                         & Qwen2.5-Omni-7B       & 44.9                          & 44.7                    & 44.9                       & 44.8                    & 45.2                       & 44.8                    & 45.0                       & 44.9                    & \textbf{45.9}              & 45.4                    & 45.4                       & 45.3                    & \textbf{45.9}             \\
MSP-Podcast              & Kimi-Audio          & 47.9                          & 47.9                    & 47.8                       & 47.9                    & 47.8                       & \textbf{48.0}           & 47.6                       & 47.9                    & 47.9                       & 47.7                    & 46.7                       & 47.7                    & 46.4                      \\
                         & Audio Flamingo 3       & 51.9                          & 51.9                    & 52.3                       & 52.2                    & 52.4                       & 51.7                    & 52.6                       & 52.5                    & 53.0                       & 51.0                    & 52.6                       & 52.9                    & \textbf{54.0}             \\ \bottomrule
\end{tabular}%
}
\caption{Agnostic injection results using different injection strategies across three $\alpha$ values. For the Mix method, we are showing the results for $\tau=0.5$.}
\label{table:injection_results_full}
\end{table*}

\paragraph{Effect of Identification Pool Size.}
Figure~\ref{fig:rvalue_example}(d) examines how many correctly answered instances per emotion are required to obtain stable ESNs. The curves plateau rapidly: a small identification pool already produces intervention effects comparable to those obtained with hundreds of instances, with larger pools yielding diminishing returns. This indicates that once the model observes a modest number of representative utterances, neuron rankings and downstream intervention behavior become largely stable. Overall, these results suggest that stable neuron identification does not require extremely large pools, enabling efficient analysis even for low-resource emotion categories.

\subsection{Activation Steering}
\paragraph{Targeted Steering.}

Table~\ref{table:ablate_results} right half shows that amplifying the same ESNs identified for deactivation yields complementary, constructive effects. Across all three LALMs, MAD and ConAct produce consistent self-steering gains (approximately $+2$ to $+3$ accuracy points) while leaving cross-steering effects largely unchanged on average, resulting in the largest self--cross gaps.
In contrast, RND is effectively neutral, and LAP/LAPE yield only small or unstable improvements, mirroring their weaker selectivity under deactivation. This symmetry between deactivation and steering strengthens the causal interpretation: neurons that are important for recognizing a target emotion can be sufficient to bias predictions toward that emotion when amplified, without broadly affecting others.

\begin{figure*}[ht!]
    \centering
    \begin{subfigure}[b]{0.165\textwidth}
        \centering
        \includegraphics[width=\linewidth]{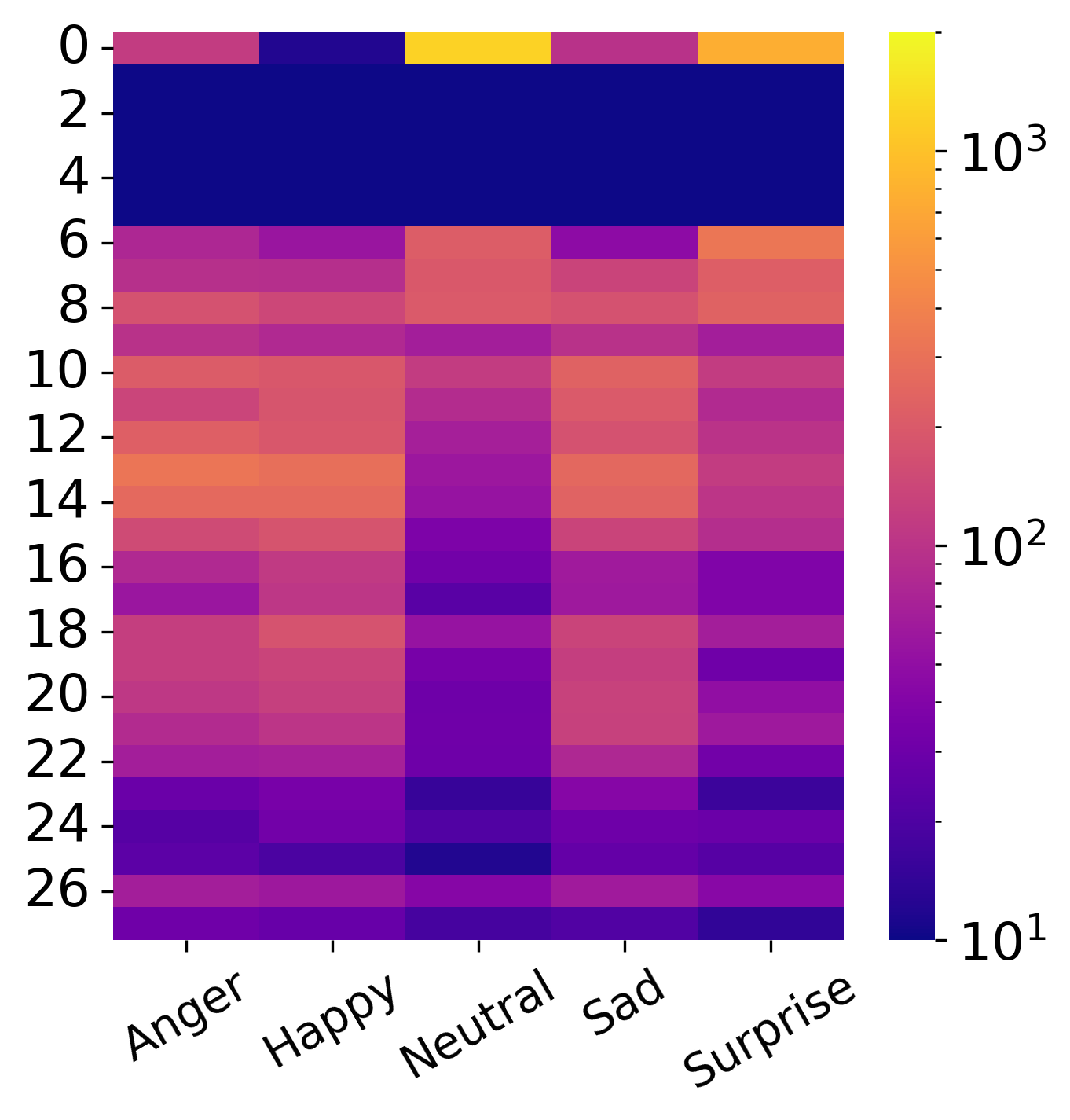}
        \caption{Qwen2.5-Omni}
    \end{subfigure}\hfill
    \begin{subfigure}[b]{0.165\textwidth}
        \centering
        \includegraphics[width=\linewidth]{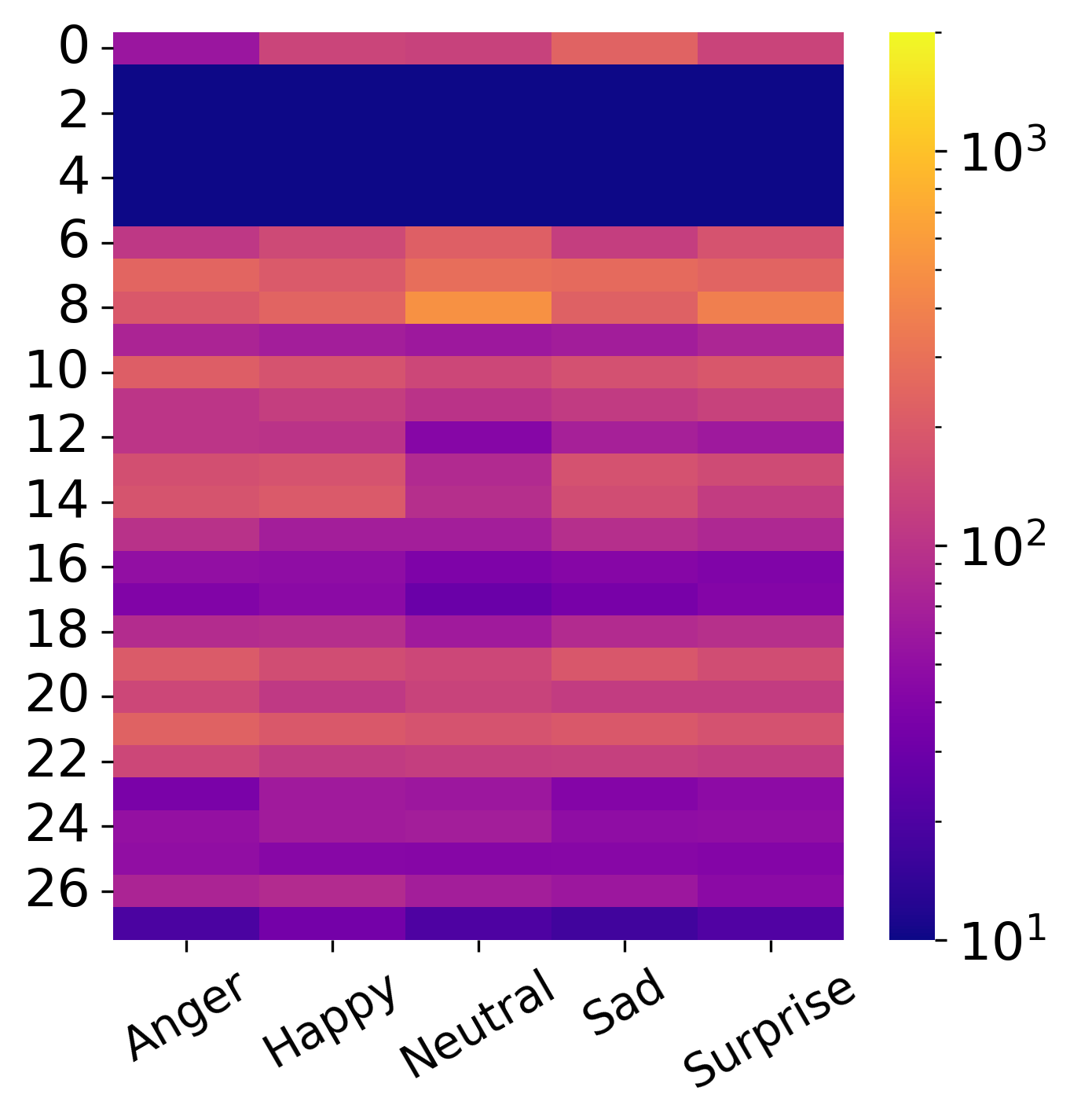}
        \caption{Kimi-Audio}
    \end{subfigure}\hfill
    \begin{subfigure}[b]{0.165\textwidth}
        \centering
        \includegraphics[width=\linewidth]{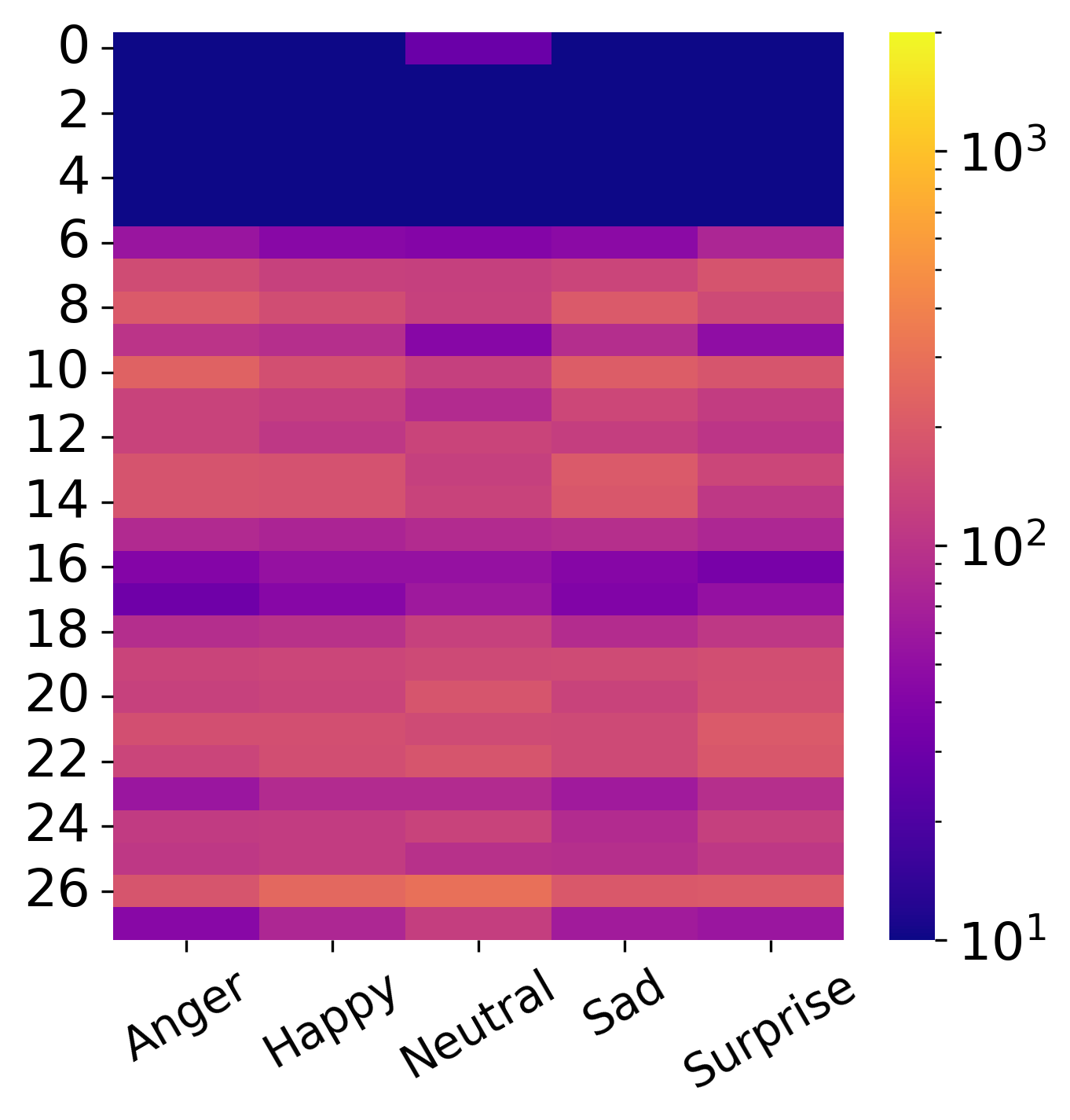}
        \caption{Audio Flamingo 3}
    \end{subfigure}\hfill
    \begin{subfigure}[b]{0.165\textwidth}
        \centering
        \includegraphics[width=\linewidth]{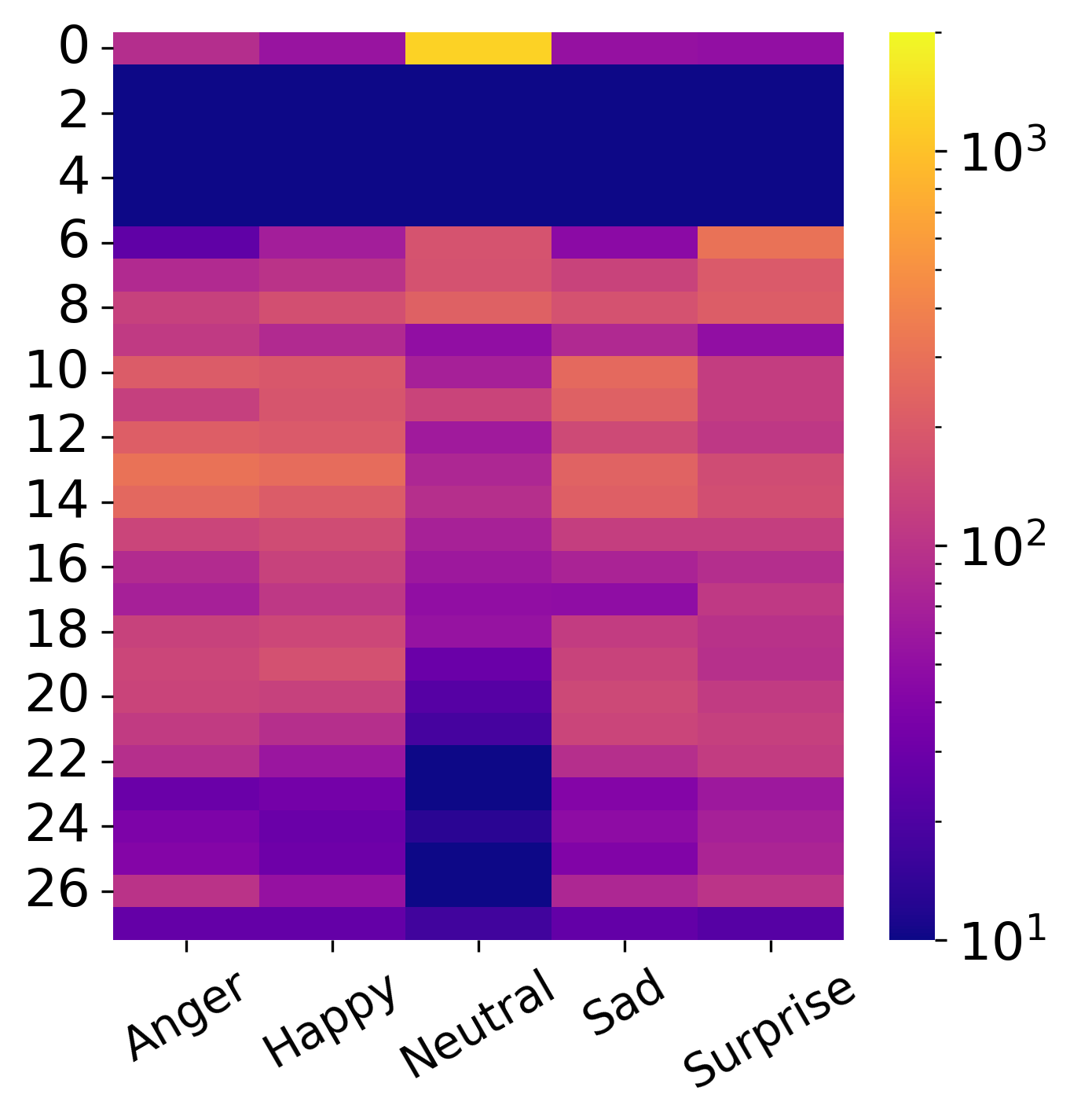}
        \caption{Qwen2.5-Omni}
    \end{subfigure}\hfill
    \begin{subfigure}[b]{0.165\textwidth}
        \centering
        \includegraphics[width=\linewidth]{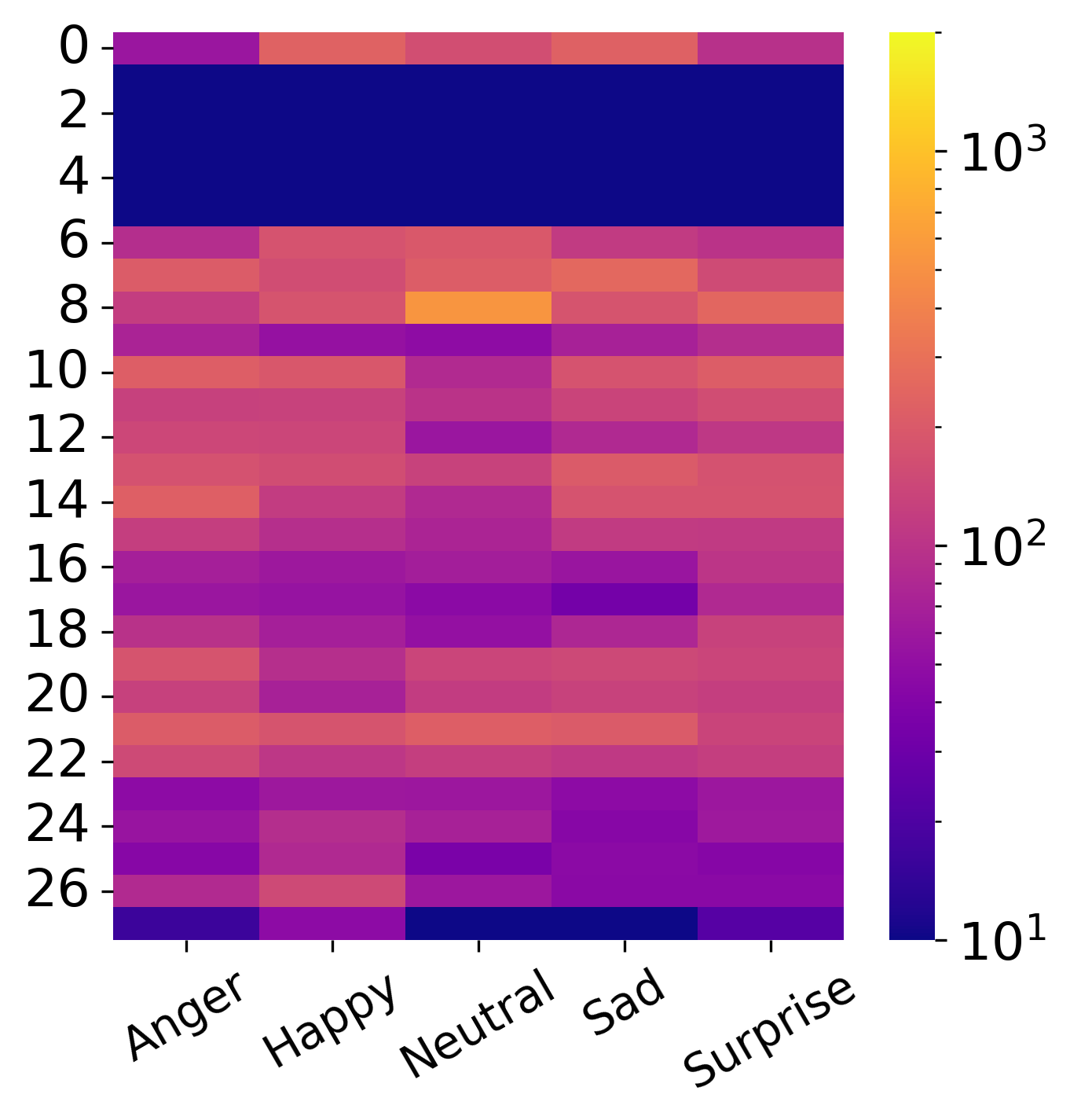}
        \caption{Kimi-Audio}
    \end{subfigure}\hfill
    \begin{subfigure}[b]{0.165\textwidth}
        \centering
        \includegraphics[width=\linewidth]{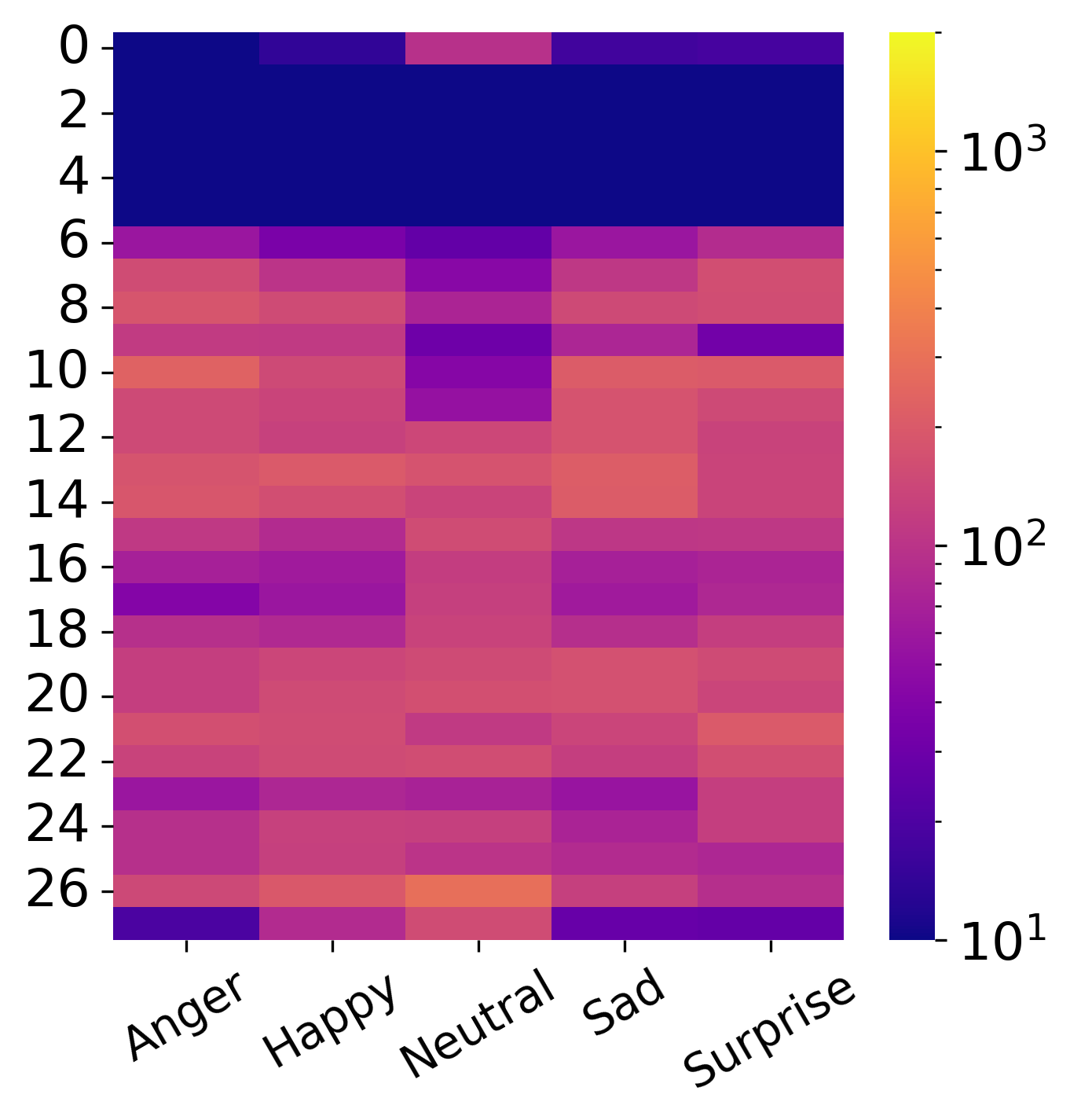}
        \caption{Audio Flamingo 3}
    \end{subfigure}
    \caption{Layer-wise distribution of identified ESNs by MAD (subfigure a, b, c) and ConAct (subfigure d, e, f). All three models have 28-layer decoders. The color is log-scaled for better readability.}
    \label{fig:layers}
\end{figure*}

Figure~\ref{fig:steer_qwen} further illustrates a strength--specificity trade-off as the steering gain $\alpha$ increases. At low $\alpha$, effects remain strongly diagonal, while larger $\alpha$ amplifies the diagonal gain but can introduce modest off-diagonal spillover. We view spillover not merely as noise, but as potential evidence that ESNs are not fully independent: sufficiently strong amplification can perturb shared downstream computation, revealing coupling (and potential competition) between affective pathways. Overall, targeted steering demonstrates that ESNs provide an actionable handle for controlled, emotion-specific behavior modulation.

\begin{figure*}[ht!]
    \centering
    \begin{subfigure}[b]{0.28\textwidth}
        \centering
        \includegraphics[width=\linewidth]{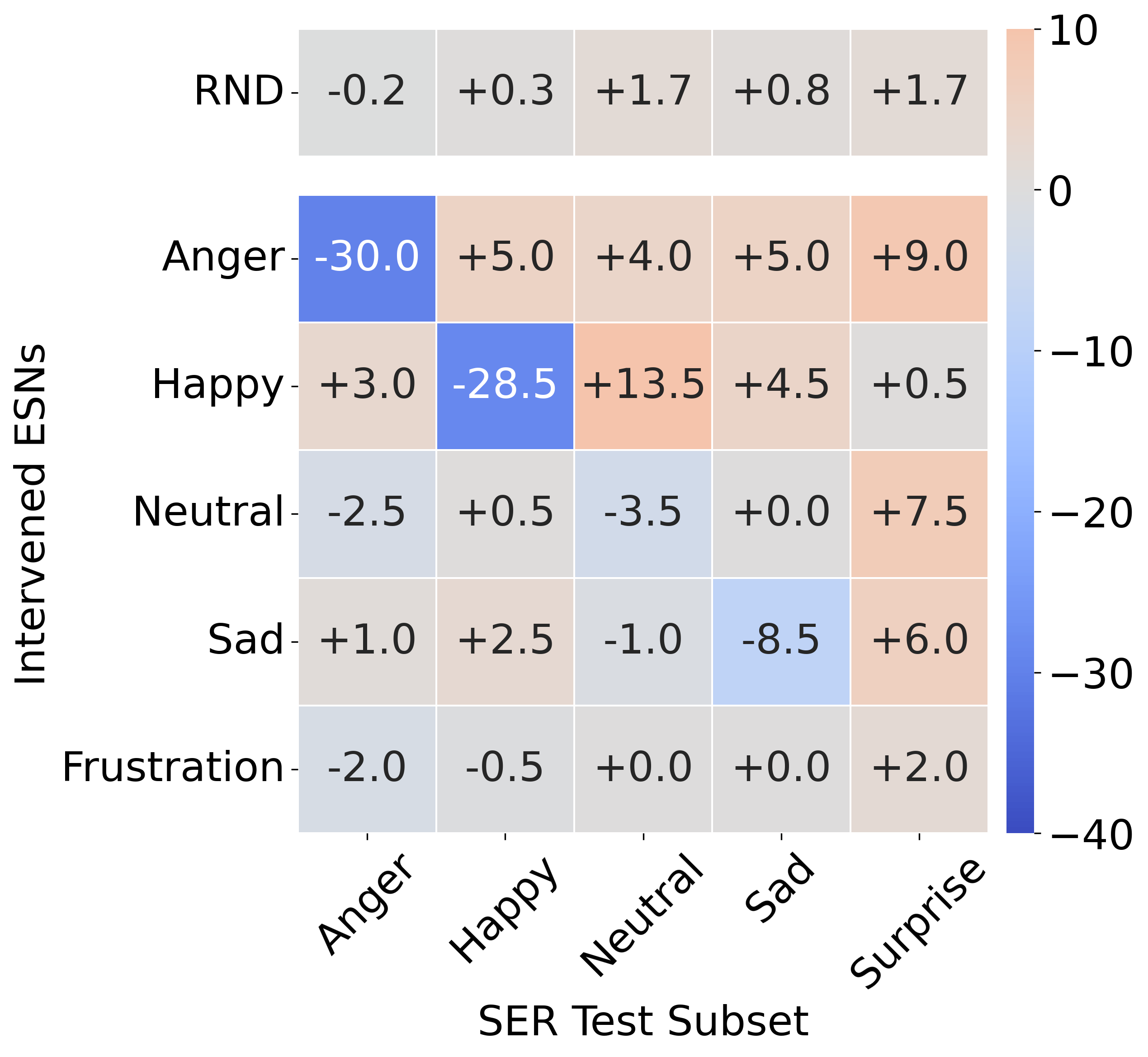}
        \caption{ESNs identified on \textsc{IEMOCAP}, tested on \textsc{MELD}}
    \end{subfigure}\hspace{0.03\textwidth}
    \begin{subfigure}[b]{0.28\textwidth}
        \centering
        \includegraphics[width=\textwidth]{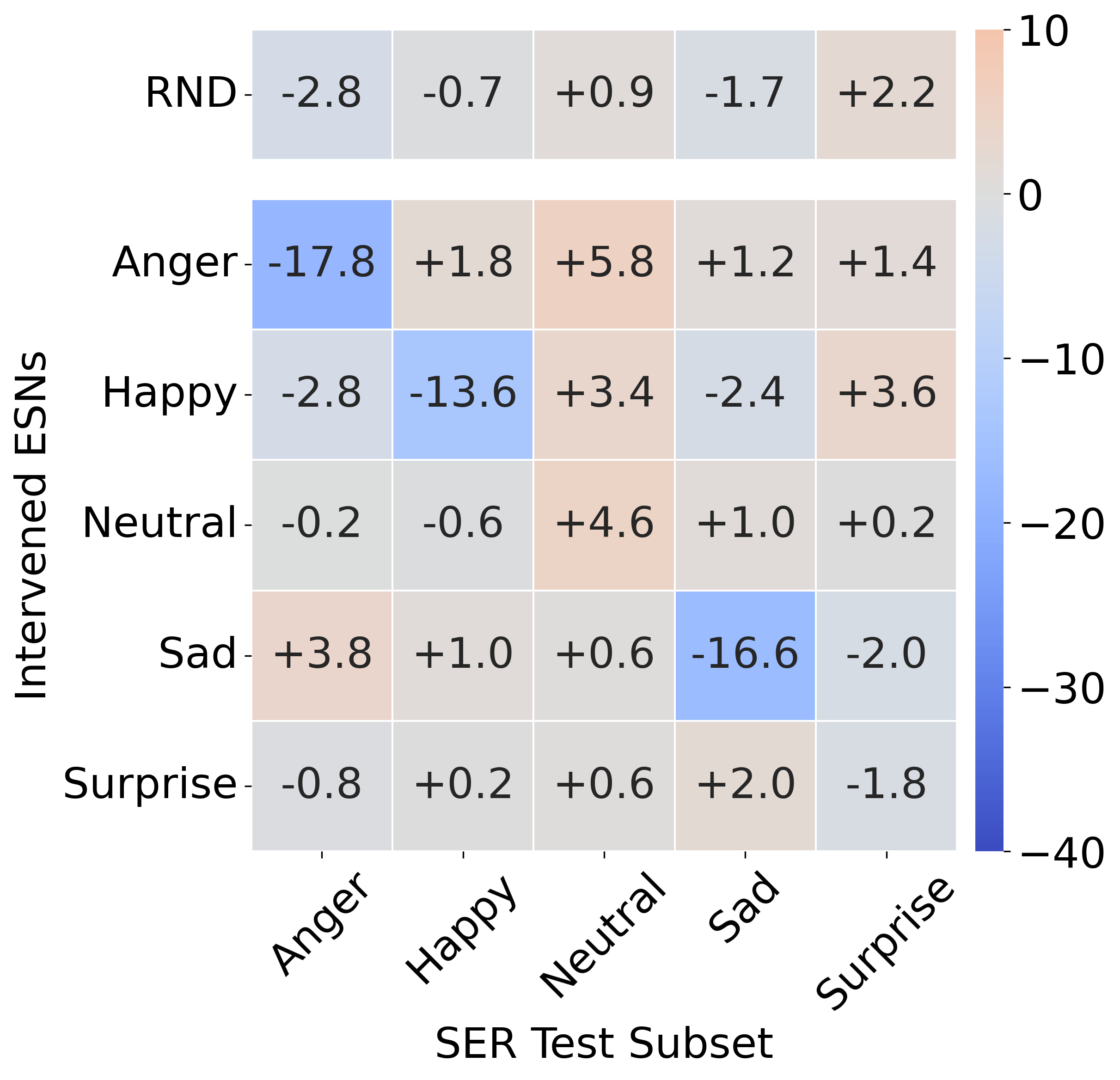}
        \caption{ESNs identified on \textsc{MELD}, tested on \textsc{MSP-Podcast}}
    \end{subfigure}\hspace{0.03\textwidth}
    \begin{subfigure}[b]{0.29\textwidth}
        \centering
        \includegraphics[width=\linewidth]{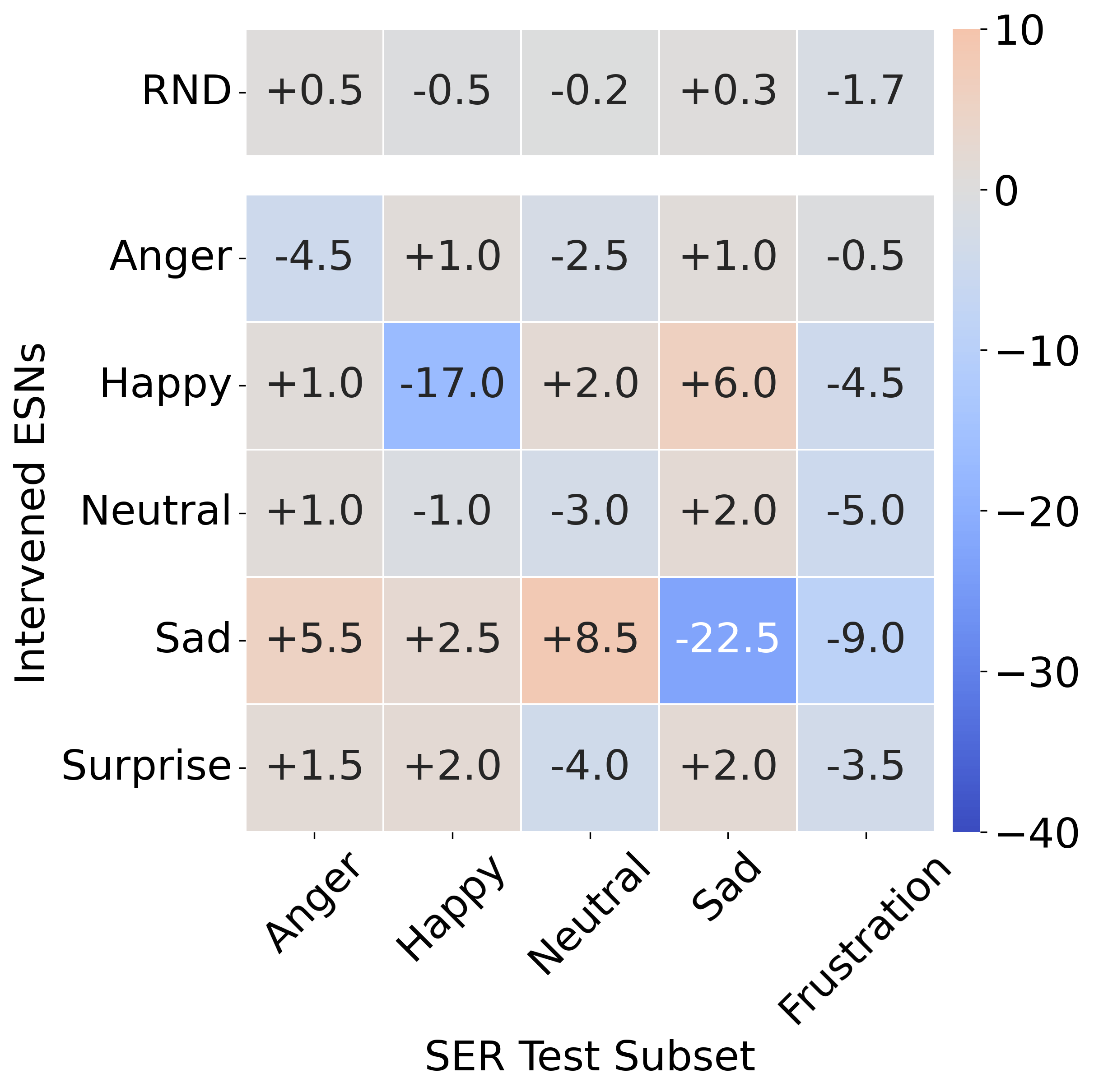}
        \caption{ESNs identified from \textsc{MSP-Podcast}, evaluated on \textsc{IEMOCAP}}
    \end{subfigure}
    \medskip
    \begin{subfigure}[b]{0.28\textwidth}
        \centering
        \includegraphics[width=\linewidth]{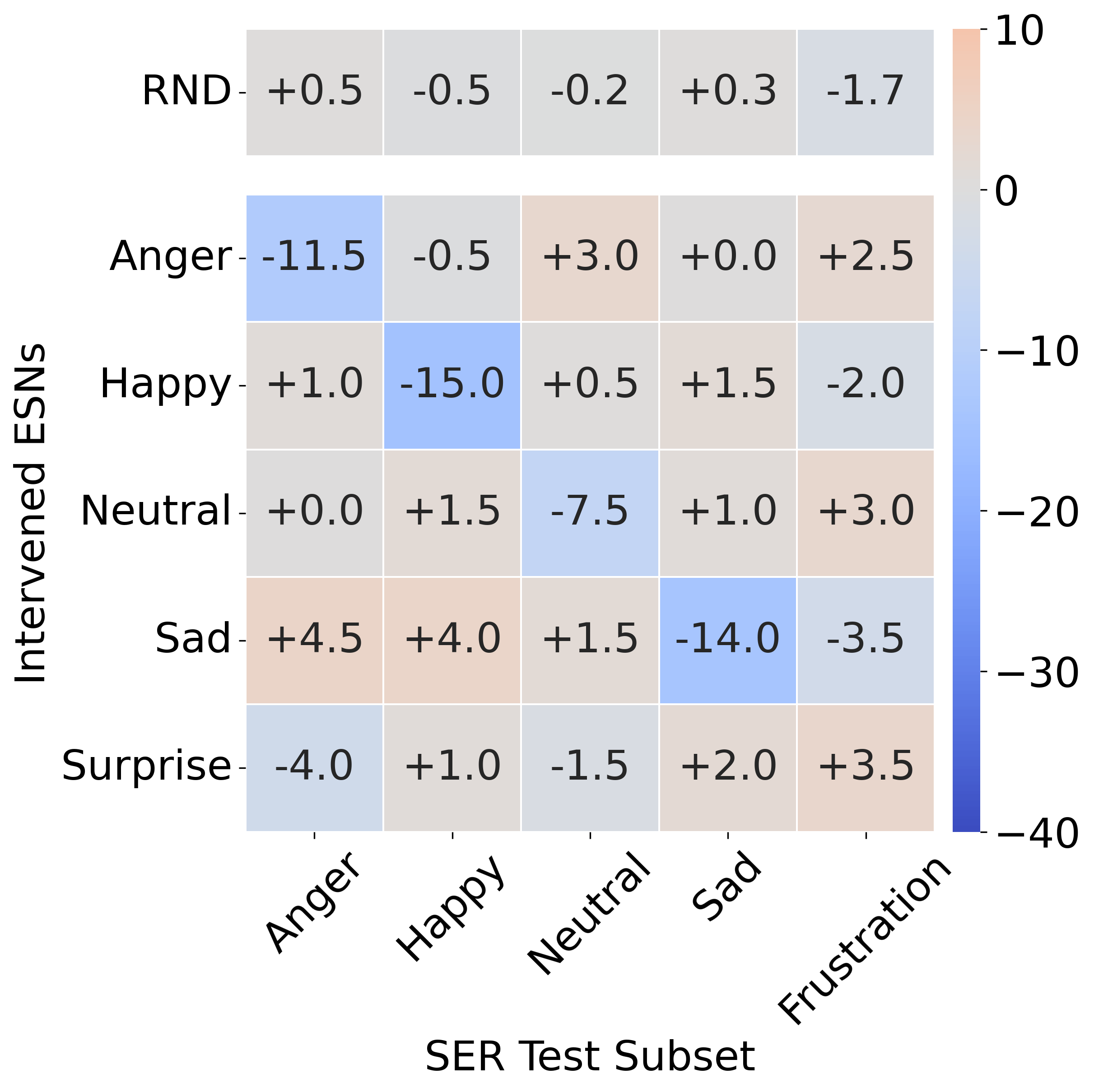}
        \caption{ESNs identified on \textsc{MELD}, tested on \textsc{IEMOCAP}}
    \end{subfigure}\hspace{0.03\textwidth}
    \begin{subfigure}[b]{0.28\textwidth}
        \centering
        \includegraphics[width=\textwidth]{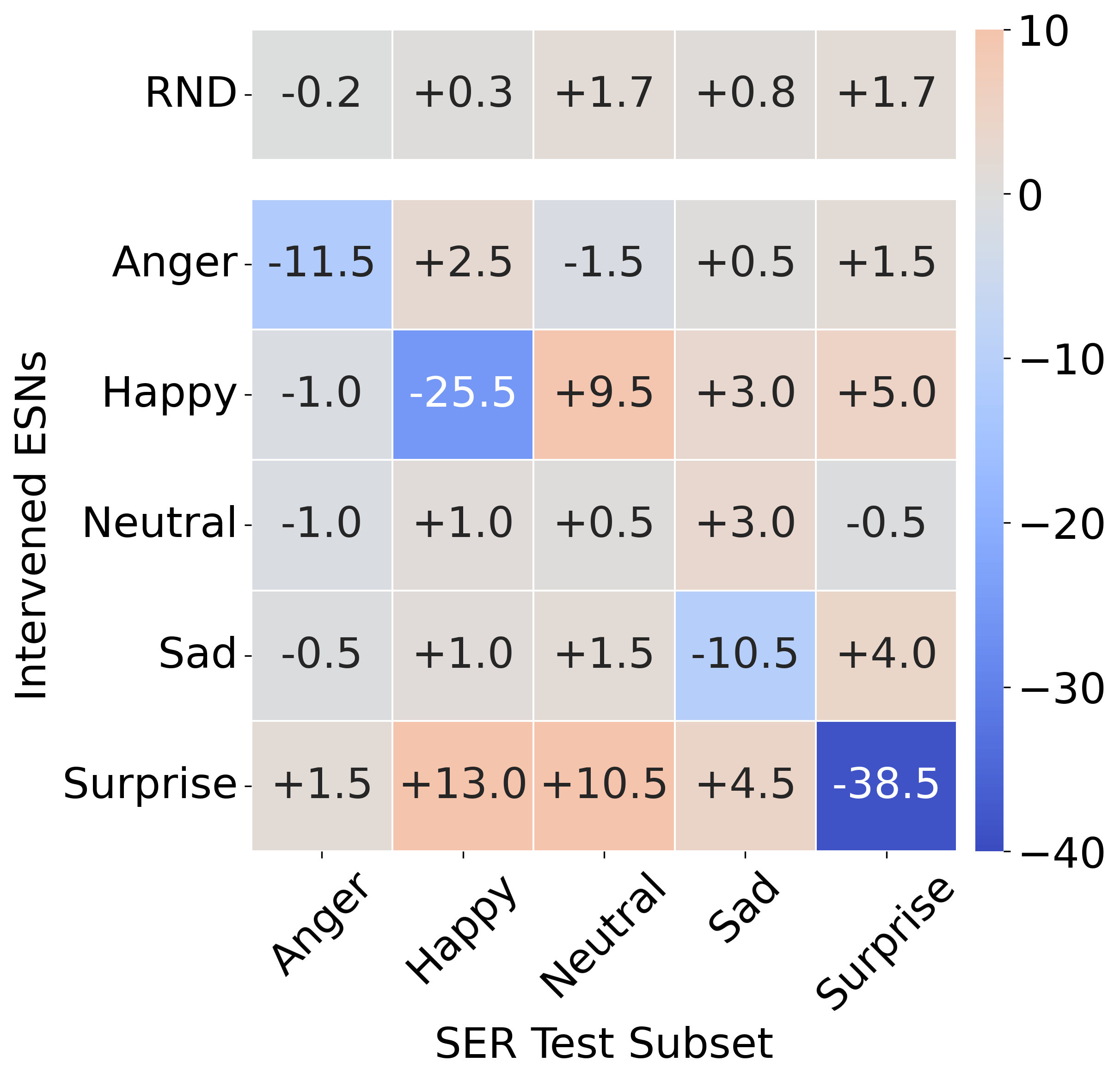}
        \caption{ESNs identified on \textsc{MSP-Podcast}, tested on \textsc{MELD}}
    \end{subfigure}\hspace{0.03\textwidth}
    \begin{subfigure}[b]{0.29\textwidth}
        \centering
        \includegraphics[width=\textwidth]{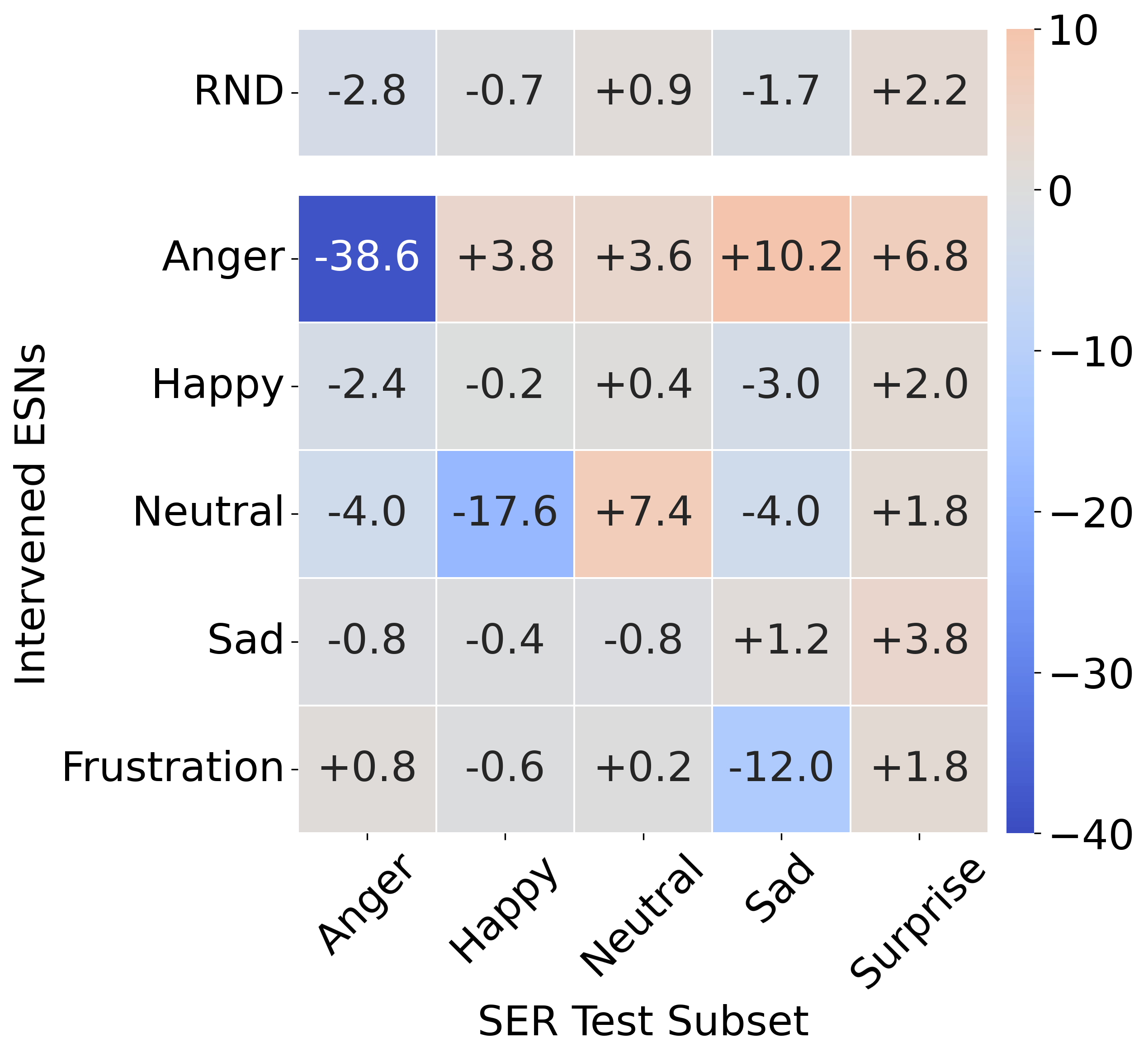}
        \caption{ESNs identified from \textsc{IEMOCAP}, evaluated on \textsc{MSP-Podcast}}
    \end{subfigure}\hspace{0.03\textwidth}
    \caption{Accuracy-change heatmaps on \textbf{cross-dataset deactivation}. The results of Qwen2.5-Omni-7B using ConAct selector are shown. Note that while all datasets contain Anger, Joy/Happiness, Neutral and Sadness; \textsc{MELD} and \textsc{MSP-Podcast} additionally share Surprise; \textsc{IEMOCAP} has Frustration.}
    \label{fig:deact_cross}
\end{figure*}

\paragraph{Agnostic Injection.}
Unlike targeted steering, agnostic injection does not condition on a source emotion. 
As summarized in Table~\ref{table:injection_results_full} provides dataset-wise results), gains are modest and model-dependent: Mix and Union improve Qwen2.5-Omni-7B (up to $+0.9$ for Mix) and Audio Flamingo 3 (up to $+1.0$ for Union), but fail to consistently benefit Kimi-Audio, where all strategies slightly underperform the unmasked baseline. In contrast to the strong and consistent targeted steering gains, this suggests that the naive Union injection over all ESNs can trigger inter-emotion interference. Concretely, 2-Pass may reinforce early mistakes by amplifying the ESNs associated with the model's first-pass prediction, while Union injects a broad ESN set that can push multiple affective directions simultaneously, reducing decisiveness when the activated units are misaligned with the true affect. Mix offers a softer compromise, but still lacks consistency across models, consistent with partial cancellation among competing affective pathways. Overall, these results indicate that basic agnostic injection is a weaker and less reliable control mechanism than targeted steering, and they hint at a non-trivial competitive structure among ESNs.

\subsection{Locality and Transferability}
\paragraph{Locality.}

Figure~\ref{fig:layers} shows the layer-wise distribution of ESNs identified by MAD and ConAct on \textsc{MSP-Podcast} across the three models. ESNs consistently cluster in the earliest layer (layer~0), early--mid layers (6--8), and later layers (19--22), with relatively sparse presence in central blocks (15--18). Both methods largely bypass these middle layers.
Interestingly, the Neutral category exhibits the strongest emotion-specific deviations, for which both methods pick unique patterns. Overall, these results indicate that the layer distribution of ESNs depends on both the identification method and the emotion category, suggesting that different affective states engage different depths of the network.

\paragraph{Cross-Dataset Transferability.}

We further investigate the dataset-independent generalization of ESNs by evaluating whether ESNs identified on one dataset remain causally effective when deactivated on another.
Figure~\ref{fig:deact_cross} showcases four source--target dataset transfers.
Across all six transfer directions, we observe recurring diagonal structure for shared emotions, indicating that many ESNs encode more dataset-robust affective computations rather than corpus-specific artifacts.
However, transfer strength is uneven and sometimes asymmetric: certain source--target pairs preserve strong self-deactivation effects, while others degrade, consistent with differences in speaking style, acoustic conditions, and annotation practices across datasets.
Among all emotions, Neutral exhibits the least stable transfer by often showing smaller or non-diagonal effects, which suggests that Neutral may rely more on dataset-dependent decision heuristics (or the absence-of-evidence boundary) than on a single portable neuron subset. These results point to a mixed but encouraging picture: ESNs show partial transfer, but their strength and selectivity depend on both the emotion category and the source--target distribution, motivating multi-dataset identification for robust control.

\section{Conclusion}
In this work, we presented a neuron-level causal study of emotion-related decisions in LALMs, and found consistent evidence that compact ESNs exist across Qwen2.5-Omni-7B, Kimi-Audio, and Audio Flamingo~3.
Across three benchmarks, we observe clear self--cross intervention signatures that are difficult to attribute to generic capacity loss: deactivating MAD/ConAct-selected ESNs produces strong emotion-specific drops while largely preserving other emotions.
Methodologically, we find that ESN identification is highly method-dependent: MAD/ConAct consistently yields more selective and stable ESN sets than LAP/LAPE or random baselines.
Actionably, amplifying the same ESNs yields reliable targeted steering gains with minimal cross-emotion spillover.
Beyond targeted control, we evaluated agnostic injection strategies and found mixed outcomes, which hints at competitive interaction among ESNs. 
We further find that ESNs stabilize with modest identification pools, exhibit non-uniform layer-wise locality and uneven, yet non-trivial cross-dataset transfer. 

Together, these findings provide evidence from causal interventions that compact, emotion-sensitive functional units exist in LALMs and that neuron-level interventions offer a practical handle for interpreting and controlling affective behavior in speech-enabled foundation models.

\section*{Limitations}
While our results consistently support the existence and controllability of ESNs in multiple LALMs, several aspects remain outside the current study's scope. Methodologically, we operationalize neuron behavior through decoder SwiGLU MLP gate activations and evaluate causality via targeted inference-time deactivation and gain-based amplification. These interventions are intentionally lightweight and comparable across architectures, but they do not fully characterize how parameterized emotion cues are distributed across other components (e.g., attention and audio–text fusion) or how multiple units compose into higher-level circuits. 
Additionally, we study transfer primarily across datasets within the SER setting. Understanding whether emotion-sensitive units generalize across tasks, such as expressive speech generation \cite{zhao2026neuronlevelemotioncontrolspeechgenerative}, and how to make steering more uniformly reliable remains an important direction for future work.
Finally, while the weaker and less stable outcomes of agnostic injection hint at inter-emotion interference among ESNs, we do not yet provide a dedicated causal decomposition of these interactions (e.g., pairwise co-steering or controlled multi-emotion activation studies); a systematic characterization of competitive vs.\ cooperative affective circuitry is an important direction for future work.

\section*{Ethical Considerations}
Our experiments are conducted on public datasets and open-sourced models, and we emphasize that our results should not be interpreted as validating emotion inference as a reliable proxy for human mental state, intent, or truthfulness. 

\section*{Acknowledgments}
This work was supported by the National Science Foundation (NSF) under CAREER Award IIS-2533652.

\bibliography{custom}

\appendix

\section{Reproducibility Details}
\label{appendix:reproducibility}

\subsection{Datasets and Models}
\label{appendix:models}

\begin{table}[H]
\centering
\resizebox{\columnwidth}{!}{%
\begin{tabular}{@{}ll@{}}
\toprule
\textbf{Models}        & \textbf{Sources}                   \\ \midrule
Qwen2.5-Omni-7B   & \url{https://huggingface.co/Qwen/Qwen2.5-Omni-7B}   \\
Kimi-Audio   & \url{https://huggingface.co/moonshotai/Kimi-Audio-7B-Instruct}  \\
Audio Flamingo 3     & \url{https://huggingface.co/nvidia/audio-flamingo-3}                  \\ \bottomrule
\end{tabular}%
}
\caption{
Sources of the evaluated models.
}
\label{table:models}
\end{table}
We employed the following three models:
\textbf{Qwen2.5-Omni-7B} \cite{xu2025qwen25omnitechnicalreport} is an end-to-end multimodal model with a streaming Thinker--Talker design.
\textbf{Kimi-Audio} \cite{kimiteam2025kimiaudiotechnicalreport} is an audio foundation model supporting audio understanding, generation, and conversational interaction.
\textbf{Audio Flamingo 3} \cite{ghosh2025audio} provides reasoning capabilities over speech, sound, and music, with support for long-form audio. 
The specific versions are listed in Table~\ref{table:models}.

Regarding the datasets, we curate balanced held-out test sets with 200 utterances per emotion for \textsc{IEMOCAP} and MELD, and 500 for MSP-Podcast, as shown in Table~\ref{table:datasets_split}. All remaining utterances are used for neuron identification, with the number of correctly answered samples per emotion controlled to ensure comparability across categories. Note that the maximum identification set sizes are determined by the lowest number of correctly answered instances per emotion per model: for example, since all models only correctly answered a little above 200 for Joy/Happiness subsets, then the maximum identification set size is set to 200.
The counts in Table 4 denote the number of correctly answered instances available for identification after excluding the held-out evaluation set (and after applying caps).

\begin{table}[ht!]
\centering
\resizebox{\columnwidth}{!}{%
\begin{tabular}{@{}lrrr@{}}
\toprule
Emotion                                                                               & Qwen2.5-Omni-7B & Kimi-Audio   & \multicolumn{1}{l}{Audio Flamingo 3} \\ \midrule
\multicolumn{4}{c}{\textit{IEMOCAP}}                                                                                                                       \\ \midrule
Anger                                                                                 & 360          & 500          & 500                                  \\
Frustration                                                                           & 500          & 500          & 500                                  \\
Joy/Happiness                                                                         & 210          & \textbf{202} & 209                                  \\
Neutral                                                                               & 500          & 500          & 500                                  \\
Sadness                                                                               & 320          & 500          & 500                                  \\ \midrule
\begin{tabular}[c]{@{}l@{}}Maximum \\ Identification Set\\ (Per Emotion)\end{tabular} & 200          & 200          & 200                                  \\
\begin{tabular}[c]{@{}l@{}}Evaluation Set\\ (Per Emotion)\end{tabular}                & 200          & 200          & 200                                  \\ \midrule
\multicolumn{4}{c}{\textit{MELD}}                                                                                                                          \\ \midrule
Anger                                                                                 & 500          & 500          & 500                                  \\
Joy/Happiness                                                                         & 500          & 500          & 500                                  \\
Neutral                                                                               & 500          & 500          & 500                                  \\
Sadness                                                                               & \textbf{228} & 326          & 404                                  \\
Surprise                                                                              & 500          & 500          & 500                                  \\ \midrule
\begin{tabular}[c]{@{}l@{}}Maximum \\ Identification Set\\ (Per Emotion)\end{tabular} & 200          & 200          & 200                                  \\
\begin{tabular}[c]{@{}l@{}}Evaluation Set\\ (Per Emotion)\end{tabular}                & 200          & 200          & 200                                  \\ \midrule
\multicolumn{4}{c}{\textit{MSP-Podcast}}                                                                                                                   \\ \midrule
Anger                                                                                 & 1000         & 1000         & 1000                                 \\
Joy/Happiness                                                                         & 1000         & 1000         & 1000                                 \\
Neutral                                                                               & 1000         & 1000         & 1000                                 \\
Sadness                                                                               & 1000         & 1000         & 1000                                 \\
Surprise                                                                              & 1000         & \textbf{810} & 1000                                 \\ \midrule
\begin{tabular}[c]{@{}l@{}}Maximum \\ Identification Set\\ (Per Emotion)\end{tabular} & 1000         & 800          & 1000                                 \\
\begin{tabular}[c]{@{}l@{}}Evaluation Set\\ (Per Emotion)\end{tabular}                & 500          & 500          & 500                                  \\ \bottomrule
\end{tabular}%
}
\caption{Correctly-answered pool size (per emotion, per model) and evaluation/identification subsampling. The per-emotion counts are (i) after holding out the evaluation set, and (ii) capped to save computation resources because we only care about the lower bounds (to determine the maximum identification set size).}
\label{table:datasets_split}
\end{table}

\subsection{Answer Normalization}
\label{appendix:normalize}

We parse model outputs into a single discrete option to make evaluation robust to minor formatting variations. Because our SER prompt requests an \emph{option number} (Appendix~\ref{appendix:prompt}), we primarily extract an integer in $\{1,\dots,|\mathcal{E}|\}$ from the generation.

Concretely, we normalize the decoded string by lowercasing, collapsing whitespace, and stripping surrounding punctuation. We then apply the following cascade:
\begin{enumerate}
    \item \textbf{Direct numeric parse.} If the output contains one or more integers in $\{1,\dots,|\mathcal{E}|\}$, we take the \emph{last} such integer as the predicted option (models may mention alternatives before concluding).
    \item \textbf{Spelled-out numbers.} If no digit is found, we map common textual forms (e.g., ``one'', ``two'') to the corresponding option index when unambiguous.
    \item \textbf{Fallback emotion-string match.} As a last resort, we match emotion names against the per-item option list (with the same normalization) and again take the last matched option if multiple appear.
\end{enumerate}
If none of the above succeeds, we mark the prediction as invalid for that item. 

\subsection{Prompt Template for SER}
\label{appendix:prompt}
\begin{lstlisting}[style=promptbox, caption={Prompt template used for SER generation, selected emotions are randomly assigned to option indices (e.g., ``1'') for each instance.}, label={lst:mcq-prompt}]
Listen to the speech clip and choose the correct emotion of the speaker:

1: {emotion 1}
2: {emotion 2}
3: {emotion 3}
4: {emotion 4}
5: {emotion 5}

Answer with the option index only.

\end{lstlisting}

\section{Agnostic Injection Methods}
\label{appendix:agnostic}
\paragraph{(1) 2-Pass Self-Consistent Injection.}
We first run the model without any intervention (Pass~1) and extract the predicted option, which we map to a predicted emotion $\hat{e}$.
In Pass~2, we attach the corresponding mask $\{I^{(m,\hat{e})}_l\}_l$ and apply the corresponding gain-based amplification. This procedure mirrors bootstrapping/self-training in that it uses the model's own first-pass decision as a pseudo-label, and reinforces it via a second-pass intervention \cite{yarowsky-1995-unsupervised,mcclosky-etal-2006-effective,NEURIPS2022_639a9a17}:
\begin{align}
\tilde g^{\text{2pass}}_{l,t} = g_{l,t} \odot s^{(\hat e)}_l(\alpha),
\\
s^{(\hat e)}_{l,n}(\alpha)=
\begin{cases}
1+\alpha, & n \in \mathcal{I}^{(m,\hat e)}_l,\\
1, & \text{otherwise.}
\end{cases}
\end{align}
We use the Pass~2 output as the final prediction. Intuitively, 2-Pass aims to make the model more \emph{self-consistent} by amplifying ESNs associated with its own inferred affect \cite{wang2023selfconsistencyimproveschainthought}.

\paragraph{(2) Mix Injection.}
Mix can be viewed as a soft, label-free compromise between no intervention and full targeted steering, guided by the model's instantaneous internal evidence; the temperature $\tau$ regulates how confidently the method concentrates on a single emotion versus spreading mass across multiple emotions \cite{NIPS2004_96f2b50b,JMLR:v23:21-0998}.

For each layer $l$ and emotion $e\in\mathcal E$, we compute an evidence score from the current gate activations by averaging over neurons in that emotion's mask and over token positions:
\begin{align}
q^{(e)}_l &= \mathbb{E}_t\Big[\mathbb{E}_{n\in I^{(m,e)}_l}\big[g_{l,t,n}\big]\Big].
\end{align}
We convert these scores into mixture weights with a temperature-controlled softmax:
\begin{align}
w^{(e)}_l &= \frac{\exp\big(q^{(e)}_l/\tau\big)}{\sum_{e'\in\mathcal E}\exp\big(q^{(e')}_l/\tau\big)},
\end{align}
where $\tau>0$ controls sharpness (smaller $\tau$ yields a more peaked distribution). Finally, we apply a per-emotion scaled gain:
\begin{align}
\tilde g^{Mix}_{l,t,n} &=
\begin{cases}
g_{l,t,n}\cdot\big(1+\alpha\,w^{(e)}_l\big), & n\in I^{(m,e)}_l,\\
g_{l,t,n}, & \text{otherwise}.
\end{cases}
\end{align}
\noindent\textbf{Overlapping masks.}
If a neuron index $n$ belongs to multiple emotion-specific sets at layer $l$ (i.e., $n \in I_{l}^{(m,e)}$ for more than one $e$), we apply the strongest multiplicative gain:
\[
\tilde{g}_{l,t,n}^{(m)} \;=\; g_{l,t,n}^{(m)} \cdot \max_{e:\, n \in I_{l}^{(m,e)}} \bigl(1 + \alpha\, w_{l}^{(e)}\bigr),
\]
and otherwise $\tilde{g}_{l,t,n}^{(m)} = g_{l,t,n}^{(m)}$.

\paragraph{(3) Union Injection.}
Union injection is a single-pass label-free baseline that amplifies all ESNs regardless of emotion identity. It corresponds to using no disambiguating routing signal (cf.\ routing vs.\ dense activation in mixture-style models) \cite{JMLR:v23:21-0998}. We first form the layer-wise union set:
\begin{align}
U_l &= \bigcup_{e\in\mathcal E} I^{(m,e)}_l,
\end{align}
and then apply the same gain to every neuron in $U_l$:
\begin{align}
\tilde g^{Union}_{l,t,n} &=
\begin{cases}
g_{l,t,n}\cdot(1+\alpha), & n\in U_l,\\
g_{l,t,n}, & \text{otherwise}.
\end{cases}
\end{align}
Compared to Mix, Union does not attempt to infer which emotion is currently active; it provides a simple way to globally boost emotion-related circuitry in one forward pass, at the cost of reduced specificity.

\onecolumn
\section{Additional Results}

\subsection{Dataset-Specific Deactivation Results}
\label{appendix:deact_results}

\begin{table*}[ht!]
\centering
\resizebox{0.7\textwidth}{!}{%
\begin{tabular}{@{}ll|rrrrr@{}}
\toprule
LALM             & Acc.$\Delta$            & RND     & LAP     & LAPE    & MAD              & ConAct               \\ \midrule
                 & Self-Deactivation       & $-$0.32 & $-$9.00  & $-$0.30  & $-$12.90         & \textbf{$-$13.50}  \\
Qwen2.5-Omni-7B & Cross-Deactivation Avg. & --      & $-$8.97 & 0.25    & $-$0.12          & \textbf{1.43}     \\
                 & Self--Cross Gap         & --      & $-$0.03 & $-$0.55 & $-$12.78         & \textbf{$-$14.92} \\ \midrule
                 & Self-Deactivation       & $-$0.62  & 0.40 & $-$1.80    & \textbf{$-$16.00}   & $-$13.00 \\
Kimi-Audio    & Cross-Deactivation Avg. & --      & $-$0.65 & $-$0.60    & $-$2.60            & \textbf{0.12}     \\
                 & Self--Cross Gap         & --      & 1.05 & $-$1.20    & $-$13.40         & $-$13.12 \\ \midrule
                 & Self-Deactivation       & $-$0.30    & $-$32.90 & $-$6.10    & \textbf{$-$14.60} & $-$13.70           \\
Audio Flamingo 3    & Cross-Deactivation Avg. & --      & $-$33.07 & $-$0.81 & $-$0.65          & \textbf{0.43}     \\
                 & Self--Cross Gap         & --      & 0.17 & $-$5.25    & $-$12.72 & \textbf{$-$14.07}           \\ \bottomrule
\end{tabular}%
}
\caption{Deactivation results on \textsc{IEMOCAP} using ESNs selected by five identification methods.}
\label{table:ablate_iemocap}
\end{table*}

\begin{table*}[ht!]
\centering
\resizebox{0.7\textwidth}{!}{%
\begin{tabular}{@{}ll|rrrrr@{}}
\toprule
LALM             & Acc.$\Delta$            & RND     & LAP      & LAPE    & MAD               & ConAct               \\ \midrule
                 & Self-Deactivation       & 0.86 & $-$4.90  & 1.90    & $-$16.90          & \textbf{$-$19.60} \\
Qwen2.5-Omni-7B & Cross-Deactivation Avg. & --      & $-$4.72  & 0.68    & 1.33              & \textbf{3.40}     \\
                 & Self--Cross Gap         & --      & $-$0.18  & 1.23    & $-$18.23          & \textbf{$-$23.00} \\ \midrule
                 & Self-Deactivation       & $-$0.62 & 0.30     & $-$1.20 & \textbf{$-$12.50} & $-$10.70 \\
Kimi-Audio    & Cross-Deactivation Avg. & --      & 0.03     & $-$1.53 & $-$0.98           & \textbf{0.10}     \\
                 & Self--Cross Gap         & --      & 0.28     & 0.32    & \textbf{$-$11.53} & $-$10.80 \\ \midrule
                 & Self-Deactivation       & $-$0.22 & $-$36.40 & $-$5.40 & $-$15.00 & \textbf{$-$15.70} \\
Audio Flamingo 3    & Cross-Deactivation Avg. & --      & $-$36.80 & $-$1.87 & $-$0.15           & \textbf{1.10}     \\
                 & Self--Cross Gap         & --      & 1.40     & $-$3.52 & $-$14.85 & \textbf{$-$16.80} \\ \bottomrule
\end{tabular}%
}
\caption{Deactivation results on \textsc{MELD}.}
\label{table:ablate_meld}
\end{table*}

\begin{table*}[ht!]
\centering
\resizebox{0.7\textwidth}{!}{%
\begin{tabular}{@{}ll|rrrrr@{}}
\toprule
LALM             & Acc.$\Delta$            & RND     & LAP      & LAPE    & MAD               & ConAct               \\ \midrule
                 & Self-Deactivation       & 0.41    & $-$8.96  & 1.52    & \textbf{$-$9.48}  & $-$7.40           \\
Qwen2.5-Omni-7B & Cross-Deactivation Avg. & --      & $-$8.29  & $-$0.81 & $-$0.65           & \textbf{0.43}     \\
                 & Self--Cross Gap         & --      & $-$0.67  & 2.33    & \textbf{$-$8.83}  & $-$7.83           \\ \midrule
                 & Self-Deactivation       & $-$0.30 & 0.12     & 0.56    & \textbf{$-$12.40} & $-$11.24          \\
Kimi-Audio    & Cross-Deactivation Avg. & --      & $-$1.04  & $-$0.63 & $-$0.24           & \textbf{1.10}     \\
                 & Self--Cross Gap         & --      & 1.16     & 1.19    & $-$12.16          & \textbf{$-$12.34} \\ \midrule
                 & Self-Deactivation       & 0.14    & \textbf{$-$34.56} & $-$7.96 & $-$15.92 & $-$14.48         \\
Audio Flamingo 3    & Cross-Deactivation Avg. & --      & $-$35.04 & $-$2.91 & $-$3.85           & \textbf{0.64}         \\
                 & Self--Cross Gap         & --      & 0.48     & $-$5.05 & $-$12.07 & \textbf{$-$15.12}         \\ \bottomrule
\end{tabular}%
}
\caption{Deactivation results on \textsc{MSP-Podcast}}.
\label{table:ablate_msp}
\end{table*}

\newpage
\subsection{Dataset-Specific Targeted Steering Results}
\label{appendix:steer_results}

\begin{table*}[ht!]
\centering
\resizebox{0.7\textwidth}{!}{%
\begin{tabular}{@{}ll|rrrrr@{}}
\toprule
LALM             & Acc.$\Delta$        & RND     & LAP           & LAPE    & MAD              & ConAct           \\ \midrule
                 & Self-Steering       & $-$0.54 & $-$0.20       & $-$0.90 & 0.90             & \textbf{2.20} \\
Qwen2.5-Omni-7B & Cross-Steering Avg. & --      & $-$0.50       & $-$1.05 & \textbf{$-$0.08} & $-$0.90       \\
                 & Self--Cross Gap     & --      & 0.30          & 0.15    & 0.98             & \textbf{3.10} \\ \midrule
                 & Self-Steering       & $-$0.38 & $-$1.50       & $-$0.10 & \textbf{2.10}    & \textbf{2.10} \\
Kimi-Audio       & Cross-Steering Avg. & --      & $-$1.10       & 0.50    & $-$0.90          & $-$1.23       \\
                 & Self--Cross Gap     & --      & $-$0.40       & 0.40    & 3.00             & \textbf{3.32} \\ \midrule
                 & Self-Steering       & 0.02    & $-$0.50       & 1.30    & 2.90             & \textbf{3.10} \\
Audio Flamingo 3 & Cross-Steering Avg. & --      & \textbf{0.00} & $-$0.65 & $-$0.68          & $-$0.53       \\
                 & Self--Cross Gap     & --      & $-$0.50       & 1.95    & 3.58             & \textbf{3.63} \\ \bottomrule
\end{tabular}%
}
\caption{Targeted steering results on \textsc{IEMOCAP}.}
\label{table:steer_iemocap}
\end{table*}

\begin{table*}[ht!]
\centering
\resizebox{0.7\textwidth}{!}{%
\begin{tabular}{@{}ll|rrrrr@{}}
\toprule
LALM             & Acc.$\Delta$        & RND     & LAP              & LAPE             & MAD              & ConAct           \\ \midrule
                 & Self-Steering       & 0.60    & 0.40             & $-$0.10          & 4.30             & \textbf{4.70} \\
Qwen2.5-Omni-7B & Cross-Steering Avg. & --      & 0.30             & 0.57             & \textbf{$-$0.05} & $-$0.45       \\
                 & Self--Cross Gap     & --      & 0.10             & $-$0.67          & 4.35             & \textbf{5.15} \\ \midrule
                 & Self-Steering       & $-$0.28 & $-$1.00          & $-$0.60          & \textbf{2.10}    & 1.40          \\
Kimi-Audio       & Cross-Steering Avg. & --      & $-$1.05          & \textbf{$-$0.15} & $-$0.35          & $-$0.62       \\
                 & Self--Cross Gap     & --      & 0.05             & $-$0.45          & \textbf{2.45}    & 2.02          \\ \midrule
                 & Self-Steering       & $-$0.36 & $-$0.20          & $-$0.20          & 2.10             & \textbf{3.00} \\
Audio Flamingo 3 & Cross-Steering Avg. & --      & \textbf{$-$0.30} & $-$0.60          & $-$0.57          & $-$1.00       \\
                 & Self--Cross Gap     & --      & 0.10             & $-$0.40          & 2.67             & \textbf{4.00} \\ \bottomrule
\end{tabular}%
}
\caption{Targeted steering results on \textsc{MELD}.}
\label{table:steer_meld}
\end{table*}

\begin{table*}[ht!]
\centering
\resizebox{0.7\textwidth}{!}{%
\begin{tabular}{@{}ll|rrrrr@{}}
\toprule
LALM             & Acc.$\Delta$        & RND     & LAP           & LAPE             & MAD           & ConAct           \\ \midrule
                 & Self-Steering       & $-$0.04 & 0.16          & 2.08             & \textbf{2.24} & 1.28          \\
Qwen2.5-Omni-7B & Cross-Steering Avg. & --      & \textbf{0.19} & 0.35             & $-$0.62       & $-$0.46       \\
                 & Self--Cross Gap     & --      & $-$0.03       & 1.73             & \textbf{2.86} & 1.74          \\ \midrule
                 & Self-Steering       & 0.10    & $-$0.48       & $-$0.44          & \textbf{2.56} & 2.32          \\
Kimi-Audio       & Cross-Steering Avg. & --      & $-$0.19       & \textbf{$-$0.09} & $-$0.72       & 0.68          \\
                 & Self--Cross Gap     & --      & $-$0.29       & $-$0.35          & \textbf{3.28} & 3.00          \\ \midrule
                 & Self-Steering       & $-$0.27 & 0.16          & 2.08             & 3.92          & \textbf{3.96} \\
Audio Flamingo 3 & Cross-Steering Avg. & --      & 0.19          & 0.35             & \textbf{0.17} & $-$0.63       \\
                 & Self--Cross Gap     & --      & $-$0.03       & 1.73             & 3.75          & \textbf{4.59} \\ \bottomrule
\end{tabular}%
}
\caption{Targeted steering results on \textsc{MSP-Podcast}.}
\label{table:steer_msp}
\end{table*}

\end{document}